\documentclass[sn-mathphys,Numbered]{sn-jnl}

\usepackage{graphicx}%
\usepackage{multirow}%
\usepackage{amsmath,amssymb,amsfonts}%
\usepackage{amsthm}%
\usepackage{mathrsfs}%
\usepackage[title]{appendix}%
\usepackage{xcolor}%
\usepackage{textcomp}%
\usepackage{manyfoot}%
\usepackage{booktabs}%
\usepackage{algorithm}%
\usepackage{algorithmic}%
\usepackage{listings}%

\theoremstyle{thmstyleone}%
\newtheorem{thm}{Theorem}

\theoremstyle{thmstyletwo}%

\theoremstyle{thmstylethree}%

\begin{document}

\title[Article Title]{IOB: Integrating Optimization Transfer and Behavior Transfer for Multi-Policy Reuse}


\author[1]{\fnm{Siyuan} \sur{Li}}\email{siyuanli@hit.edu.cn}

\author[2]{\fnm{Hao} \sur{Li}}\email{li.hao@mail.nwpu.edu.cn}

\author[3]{\fnm{Jin} \sur{Zhang}}\email{jin-zhan20@mails.tsinghua.edu.cn}


\author[2]{\fnm{Zhen} \sur{Wang}}\email{zhenwang0@gmail.com}

\author[1]{\fnm{Peng} \sur{Liu}}\email{pengliu@hit.edu.cn}

\author*[3]{\fnm{Chongjie} \sur{Zhang}}\email{chongjie@wustl.edu}

\affil[1]{\orgdiv{Faculty of Computing}, \orgname{Harbin Institute of Technology}, \orgaddress{\city{Harbin}, \postcode{150001}, \country{China}}}

\affil[2]{\orgdiv{School of Artificial Intelligence, Optics and Electronics (iOPEN)}, \orgname{Northwestern Polytechnical University}, \orgaddress{\city{Xi'an}, \postcode{710072}, \country{China}}}

\affil*[3]{\orgdiv{McKelvey School of Engineering}, \orgname{Washington University in St. Louis}, \orgaddress{\city{St. Louis}, \postcode{63130}, \country{United States}}}


\abstract{
Humans have the ability to reuse previously learned policies to solve new tasks quickly, and reinforcement learning (RL) agents can do the same by transferring knowledge from source policies to a related target task. Transfer RL methods can reshape the policy optimization objective (optimization transfer) or influence the behavior policy (behavior transfer) using source policies. However, selecting the appropriate source policy with limited samples to guide target policy learning has been a challenge. Previous methods introduce additional components, such as hierarchical policies or estimations of source policies' value functions, which can lead to non-stationary policy optimization or heavy sampling costs, diminishing transfer effectiveness. To address this challenge, we propose a novel transfer RL method that selects the source policy without training extra components. Our method utilizes the Q function in the actor-critic framework to guide policy selection, choosing the source policy with the largest one-step improvement over the current target policy. We integrate optimization transfer and behavior transfer (IOB) by regularizing the learned policy to mimic the guidance policy and combining them as the behavior policy. This integration significantly enhances transfer effectiveness, surpasses state-of-the-art transfer RL baselines in benchmark tasks, and improves final performance and knowledge transferability in continual learning scenarios. Additionally, we show that our optimization transfer technique is guaranteed to improve target policy learning.
}




\maketitle

\section{Introduction}\label{sec1}
Trough transferring knowledge from previous policies, humans can learn to solve related new tasks quickly \cite{guberman1991learning}. However, current deep reinforcement learning (RL) agents lack this knowledge transfer ability \cite{silver2017mastering,vinyals2019grandmaster,ceron2021revisiting}, which results in inefficient learning. To address this problem, a large number of research works investigate the multi-policy reuse problem in deep RL: how to efficiently reuse the knowledge from multiple source policies to speed up the learning in a target task \cite{fernandez2006probabilistic,barreto2018transfer,li2019hierarchical,yang2020efficient,zhang2022cup}.

To achieve efficient knowledge transfer in RL, the first problem is how to use the source knowledge to influence the learning process in the target task. As there are two major parts in RL: collecting samples and optimizing policies with the collected samples, previous transfer RL works improve the learning efficiency in the target task by either utilizing the source policies to affect the behavior policy of the agent \cite{fernandez2006probabilistic, li2018context, li2019hierarchical}, which we name as \textit{behavior transfer}, or reusing the source policies to shape the optimization objective of the target policy \cite{zhang2022cup, acteach, barreto2018transfer}, which we name as \textit{optimization transfer}. Conducting behavior transfer and optimization transfer is challenging, since there are multiple source policies in the given policy set, and a proper source policy needs to be selected from this set to guide the target policy learning at an early learning stage.  

Existing research works learn to select source policies by introducing additional components, such as hierarchical high-level policies over the source policies \cite{li2018context,li2019hierarchical,yang2020efficient}, or estimating the value functions of the source policies on the target task \cite{barreto2017successor,barreto2018transfer,cheng2020policy}. However, training these additional components significantly harms the transfer effectiveness, as hierarchical policy structures induce a non-stationary issue for policy optimization  \cite{pateria2021hierarchical}, and estimating the value functions for each source policy is with high sampling cost and computationally expensive. To accomplish efficient transfer without training any additional components, we propose a novel transfer RL method, which employs the value function in the actor-critic framework \cite{lillicrap2016continuous,fujimoto2018addressing,haarnoja2018soft2} to select the guidance policy from the source policy set, and then uses the selected guidance policy to conduct the transfer. The proposed approach Integrates Optimization transfer and Behavior transfer, which is dubbed as IOB.
By inferring the Q function, IOB chooses the source policy that has the largest one-step improvement over the currently learned target policy as the guidance policy. In the policy optimization process, IOB regularizes the target policy to imitate the guidance policy. During the interaction with the environment, the guidance policy and the learned target policy are combined together to form a behavior policy to enable more efficient data collection.

The advantages of the IOB approach are as follows. (i) The one-step improvement can be estimated by querying the Q function and no additional components are needed to be trained. (ii) IOB seamlessly combines optimization transfer and behavior transfer, which accelerates the learning in the target task to a maximum extent. (iii) IOB is conceptually simple and easy to implement, as it introduces very few hyper-parameters to the backbone algorithm. (iv) The optimization transfer in IOB is theoretically guaranteed to improve the target policy learning process.  (v) IOB can be naturally integrated with existing continual RL methods to efficiently construct agents with multi-task ability. 

To evaluate the proposed method, we compare it with state-of-the-art transfer RL methods on the Meta-World benchmark \cite{yu2020meta}. Experiment results demonstrate that our method significantly outperforms the baseline methods and achieves the largest \textit{forward transfer}. Then, we visualize the guidance policy selection process to explain the reason why the proposed method works. Next, we perform several ablation studies to analyze the influence of the components of IOB on transfer performance. Finally, we demonstrate that the proposed method could be applied to a continual learning setting, where we combine IOB with a continual learning approach. Experiment results show that IOB boosts the transfer performance of the backbone continual learning approach while maintaining its stability. 

We note that a shorter conference version of this paper appeared in \cite{zhang2022cup}. Our initial conference paper has not  introduced behavior transfer. This manuscript further promotes the transfer ability of the agent by integrating behavior transfer and optimization transfer. As the guidance policy selection relies on the Q function, this manuscript proposes to increase the accuracy of Q function with ensemble learning. Furthermore, we add a continual learning experiment to better demonstrate the scope of the proposed method.  

In the remainder of this paper, we start by reviewing the background knowledge and describing the problem formulation. After that, we review the related work about multi-policy reuse. Next, we present the proposed approach followed by experiment results comparing our approach with the state-of-the-art baselines. Finally, we conclude and outline the directions for future research. 

\section{Related Works}\label{sec2}

The learning inefficiency of deep RL approaches restricts their applications to more real-world problems, and transfer learning methods have long been recognized as an effective way to improve the efficiency of the deep RL approaches \cite{zhu2020transfer, parisotto2015actor, hou2017evolutionary, laroche2017transfer}. Here we roughly classify deep transfer RL algorithms into three categories. The algorithms in the first category mainly utilize the source knowledge to reshape the optimization objectives of the benchmark RL methods, which we refer to as the \textit{optimization transfer} methods. The second category focuses on transferring the behavior of the source policies to facilitate the exploration process in the target task, which we refer to as \textit{behavior transfer} methods. The last category is devoted to transferring the parameters of the source policy networks to target policy learning, which we call as \textit{parameter transfer} methods. These three categories of methods aim to solve the policy optimization, data collection and parameter initialization challenges in deep RL respectively. 

\textbf{Optimization transfer.} The optimization transfer methods employ source knowledge to accelerate the optimization process of the target policy. The AC-Teach method \cite{acteach} uses the value estimation of the source policies to improve the target policy optimization in a Bayesian manner. Similarly, \cite{barreto2017successor} and \cite{barreto2018transfer} propose to aggregate the source policies by choosing the policy with the largest Q value at each state. These two methods assume that the source and target tasks share the same dynamics, so that they use the successor features \cite{JMLRv2119-060} to mitigate the computation cost brought by estimating the Q functions for all the source policies. In contrast, our method only estimates the Q function of the target policy, which is more computationally efficient. Besides, our method works in a more general setting, allowing different dynamic functions for the source and target. The MAMBA method \cite{cheng2020policy} forms a new baseline function by aggregating the value functions of the source policies, and then guides the policy search by improving the policy over the baseline function. The MULTIPOLAR method \cite{barekatain2021multipolar} learns a weighted sum over the actions of the source policies, and learns an auxiliary network to predict the residuals around the aggregated actions. Compared with these previous methods, our method does not require training any additional components, which is both computationally efficient and sampling efficient. 

\textbf{Behavior transfer.} The behavior transfer methods aim at improving the exploration efficiency in the target task with the given source policies. A series of works propose to combine the source policies with random policies probabilistically to achieve more efficient exploration \cite{fernandez2006probabilistic, li2018optimal,gimelfarb2021contextual}. As the combination manner is not learned, these methods cannot guarantee to perform better than learning without the source knowledge. To accomplish a more effective policy combination, the following works \cite{li2018context, li2019hierarchical,yang2020efficient,yang2021hierarchical} propose a hierarchical policy structure to reuse the given policies, where a high-level policy is learned to select which source policy should be executed at the current state.  Although the policy combination is more accurate under the hierarchical structure, the simultaneous learning of multi-level policies suffer from the non-stationary issue \cite{pateria2021hierarchical}. In this work, we conduct behavior transfer without the hierarchical policy structure. Instead, the source policies and the target policy are combined under the guidance of the learned critic. 

\textbf{Parameter transfer.}
The parameter transfer methods initialize the neural networks for the target task with the parameters learned in the source tasks. When there are multiple source tasks, it is a challenging problem which source policy to transfer from, since the parameter initialization needs to be conducted before the learning in the new target task starts.  Some works propose a progressive network structure which connects these source neural networks with lateral connections \cite{rusu2016progressive, berseth2018progressive}, and then use the progressive network as the initialization in the target task. These methods may not be scalable when the number of the source tasks grow too large.  For better scalablity, some following works propose to prune the source networks or distill the source networks to a smaller network and then reuse the parameters \cite{schwarz2018progress, mallya2018packnet, teh2017distral}. Note that our method and the parameter transfer methods are orthogonal, and could be easily combined together. In section \ref{continual}, we combine the proposed method with a parameter transfer method based on pruning called PackNet \cite{mallya2018packnet}, and evaluate it in a continual learning setting. 

\section{Preliminaries and Problem Formulation}\label{sec3}
The environment in RL is formulated as a Markov Decision Process (MDP), and the MDP is defined by a tuple $(\mathcal{S},\mathcal{A},p,r,\gamma)$, where $\mathcal{S}$ is a state space, $\mathcal{A}$ is an action space, $p(s'|s,a)$ is an unknown transition function, $r(s,a,s')$ is a reward function, and $\gamma\in[0, 1)$ is a discount factor. The objective of RL is to learn a policy $\pi(a|s)$ that could maximize the expected discounted return: $R(\pi)=\mathbb{E}_{\pi}\left[\sum_{t=0}^{\infty}\gamma^tr(s_t,a_t,s_{t+1})|a_t\sim\pi(\cdot|s),s_0\right]$.

While the proposed approach could be easily integrated with off-policy actor-critic methods,  
 in the following sections, we mainly present how it could be combined with the Soft Actor-Critic (SAC) algorithm \cite{haarnoja2018soft2}. By removing the entropy term in SAC, the proposed approach can be applied to other off-policy actor-critic RL methods as well, such as Deep Deterministic Policy Gradient (DDPG) \cite{lillicrap2016continuous} and Twin-Delayed DDPG (TD3) \cite{fujimoto2018addressing}.  Next, we introduce some preliminary knowledge about the SAC method.

 \textbf{SAC:} The SAC method \cite{haarnoja2018soft} introduced an additional function approximator for the value function, but later found it to be unnecessary\cite{haarnoja2018soft2}. In this paper, the soft Q function and soft V function of a policy $\pi$ in SAC are defined as:
\begin{equation}
\begin{aligned}
    &Q_{\pi}(s,a)=r(s,a)+\gamma\mathbb{E}_{s' \sim p(\cdot|s,a)}\left[V_{\pi}(s)\right],\\
    &V_{\pi}(s)=\mathbb{E}_{a \sim \pi(\cdot|s)}\left[Q_{\pi}(s,a)-\alpha\log\pi(a|s)\right],
    \end{aligned}
\end{equation}
where $\alpha>0$ is the entropy weight, and the loss functions of SAC are defined as:
\begin{equation}
\begin{split}
  &L_{critic}(Q_{\theta})=\mathbb{E}_{(s,a,r,s') \sim \mathcal{D}}\left[(Q_{\theta}(s,a)-(r+\gamma{V_{\overline{\theta}}}(s'))\right]^2, \\
  &L_{actor}(\pi_{\phi})=\mathbb{E}_{s \sim \mathcal{D}}\left[\mathbb{E}_{a \sim \pi_{\phi}(\cdot|s)}\left[\alpha \log \pi_{\phi}(a|s)-Q_{\theta}(s,a)\right]\right],   \\
  &L_{entropy}(\alpha)=\mathbb{E}_{s \sim \mathcal{D}}\left[\mathbb{E}_{a \sim \pi_{\phi}(\cdot|s)}\left[-\alpha \log \pi_{\phi}(a|s)-\alpha \overline{\mathcal{H}}\right]\right],
\end{split}
\label{sacloss2}
\end{equation}
 where $\mathcal{D}$ is the replay buffer, $\overline{\mathcal{H}}$ is a hyper-parameter representing the target entropy, $\theta$ and $\phi$ are network parameters, $\overline{\theta}$ denote the parameters of the target network, and ${V_{\overline{\theta}}}(s)=\mathbb{E}_{a \sim \pi(a|s)}[Q_{\overline{\theta}}(s,a)-\alpha \log \pi(a|s)]$ is the target soft value function.

 Based on the SAC method, we define the \textit{soft expected advantage} of action probability distribution $\pi_i(\cdot|s)$ over policy $\pi_j$ at state $s$ as:
 \begin{equation}
    Adv_{\pi_j}(s,\pi_i)=\mathbb{E}_{a \sim \pi_i(\cdot|s)}\left[Q_{\pi_j}(s,a)-\alpha\log\pi_i(a|s)-V_{\pi_j}(s)\right].
    \label{ea}
\end{equation}
$Adv_{\pi_j}(s,\pi_i)$ measures the one-step performance improvement brought by following $\pi_i$ instead of $\pi_j$ at state $s$, and following $\pi_j$ afterwards.

\textbf{Problem Formulation:} Multi-policy reuse focuses on learning the policy for a target MDP $M$ with fewer samples through transferring knowledge from a set of source policies $\{\pi_1,\pi_2,...,\pi_n\}$.  We denote the target policy learned on $M$ as $\pi_{tar}$, and its corresponding soft Q function as $Q_{\pi_{tar}}$. In this paper, we assume that the source policies and the target policy share the same state and action spaces. This assumption generally holds for the tasks with homogeneous agents, e.g., one robot manipulating different objects, or one robot with the same reception field navigating in different environments. 

\section{Method}\label{sec4}

There are two prominent components that significantly affect the efficiency of RL: the policy optimization objective and the way of collecting samples.
Although the goal of most RL methods is to maximize the expected return, it remains an unsolved problem how to use the source knowledge to shape the policy optimization objective, so that achieving this goal costs fewer stochastic gradient descent iterations. Beyond that, the way of collecting samples (behavior policy) also plays a crucial role to improve learning efficiency, as the behavior policy  determines the quality of the training data of the neural networks. 
In this paper, we propose a novel transfer RL approach, which aims at improving the learning efficiency in the target task from both the optimization and behavior perspectives using a source policy set. 

\begin{figure*}[htbp]
\centering
\includegraphics[width=5.in]{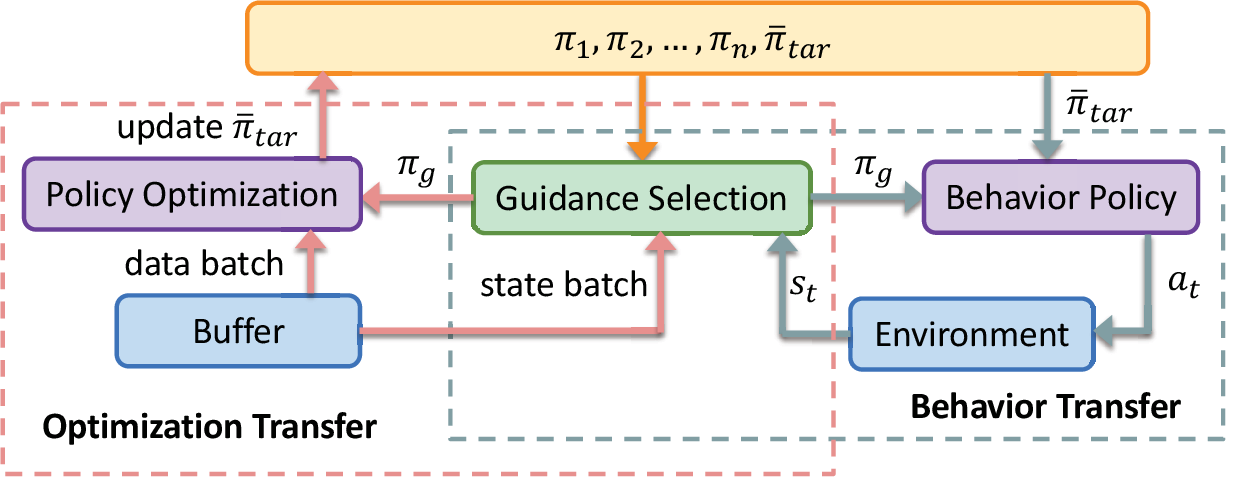}
\caption{The overall framework of the proposed approach.}
\label{fig:frame}
\end{figure*}

Figure \ref{fig:frame} visualizes the overall framework of the proposed approach, Integrating Optimization transfer and Behavior transfer for multi-policy reuse (IOB). The left dash box contains the flowchart of \textit{optimization transfer}, which is described in Section \ref{opt}, and the right dash box illustrates the \textit{behavior transfer} process, which is presented in Section \ref{beh}. Then, in Section \ref{all}, we elaborate on how those two types of transfer are combined together to thoroughly boost the target task learning efficiency. Furthermore, in Section \ref{theory}, we analyze the proposed method from a theoretical perspective and prove that even with an approximated Q function, the target policy is guaranteed to be improved monotonically with the proposed optimization transfer technique. Finally, in Section \ref{crl}, We have integrated IOB with the continual learning method to propose a novel continual RL approach highlighting transfer capabilities.

\subsection{Optimization Transfer}
\label{opt}
To achieve positive transfer from the optimization perspective, IOB utilizes action distributions output by the source policies to guide the learning of the target policy. Specifically, at state $s$, the agent has access to a set of candidate action distributions output by $n+1$ policies, including $n$ source policies and a hard copy of the target policy,  $\overline{\pi}_{tar}$:
\begin{equation}
\Pi^{s}=\{\pi_1(\cdot |s),\pi_2(\cdot |s),...,\pi_n(\cdot |s),\overline{\pi}_{tar}(\cdot |s)\}. 
\label{pis}
\end{equation}
From this candidate set, IOB selects a guidance policy $\pi_{g}$ which has the largest one-step improvement over the current target policy:
\begin{equation}
\begin{aligned}
    \pi_{g}(\cdot |s)&=\mathop{\arg\max}\limits_{\pi(\cdot |s) \in \Pi^s}Adv_{\pi_{tar}}(s,\pi) \\
    &=\mathop{\arg\max}\limits_{\pi(\cdot |s)\in \Pi^{s}}\mathbb{E}_{a \sim\pi(\cdot |s)}\left[{Q}_{\pi_{tar}}(s,a)-\alpha\log\pi(a|s)\right].
    \label{arrgrgate}
\end{aligned}
\end{equation}
The second equation holds as adding $V_{\pi_{tar}}(s)$, to all soft expected advantages does not affect the result of the $\mathop{\arg\max}$ operator. Note that the guidance policy selection is conditioned on states, and Equation \eqref{arrgrgate} indicates that the action output by the guidance policy $\pi_{g}$ at state $s$ is at least no worse than the current target policy $\pi_{tar}$ in terms of the expected return estimated by the Q value. Possibly the guidance policy is better than the current target policy, if the source policy set contains one or more policies similar to the optimal target policy. As we have obtained a guidance policy, the next question is how we could use it to guide  the target policy update. Since the guidance policy  $\pi_{g}$ may accomplish larger returns than the current target policy, we propose to regularize the target policy $\pi_{tar}$ to imitate the guidance policy $\pi_{g}$ selected from the candidate set $\Pi^s$ before this update, and minimize the following loss function:
\begin{equation}
\begin{aligned}
L_{\pi}(\pi_{tar})&=L_{actor}(\pi_{tar})\\
&+\mathbb{E}_{s \sim \mathcal{D}} \left[\beta_s D_{KL}\left(\pi_{tar}\left(\cdot|s\right)||{\pi_{g}}\left(\cdot|s\right)\right)\right],
\label{reg}
\end{aligned}
\end{equation}
where $L_{actor}$ is the original actor loss defined in Equation \eqref{sacloss2}, and $\beta_s > 0$ is a hyper-parameter controlling the weight of the regularization. The training data for the regularization is also sampled from the replay buffer $D$, the same as that of $L_{actor}$. After each update of the target policy $\pi_{tar}$, the corresponding policy in the candidate policy set is synchronized immediately.  

Since the Q value is critical to the guidance policy selection in Equation \eqref{arrgrgate}, we need a relatively accurate estimation of the Q value, so that the guidance policy could be beneficial to the target policy optimization. As the value function learning in RL often suffers from the over-estimation issue \cite{lan2020maxmin, kuznetsov2020controlling}, we propose to apply the following critic ensemble technique to attain a more accurate Q function:
\begin{equation}
    {Q}_{\pi_{tar}}(s,a) = \min_{k \in [1..K]} Q_{\theta_k}(s, a),
\end{equation}
where $\theta_k$ denotes the parameters of the $k$-th Q network. All the Q networks are independently initialized and trained. Limited to the computation resource, also balancing overestimation and underestimation of those Q functions, the total number for the Q-networks, $K$, is set as $4$ in the experiment section. 

\subsection{Behavior Transfer}
\label{beh}
The regularization in Section \ref{opt} enables faster policy learning with the training data sampled from the replay buffer, and another problem is how to fill the replay buffer with high-quality data with large returns. To solve this challenge, we propose  to further employ the guidance policy to improve the behavior policy. 
As most off-policy algorithms could only afford a slight degree of \textit{off-policyness} \cite{zhang2017deeper,fedus2020revisiting}, the proposed approach probabilistically combines the guidance policy and the learned target policy as the behavior policy in an $\epsilon$-greedy manner.

\begin{algorithm}[H]
\caption{Behavior-$\pi(\epsilon, s_t, \Pi_s, \overline{\pi}_{tar}, \{Q_{\theta_k}|k \in{1..K}\})$}\label{alg:alg1}
\begin{algorithmic}[1]
\IF{$random() <\epsilon$}
\STATE{$\pi_{b} \leftarrow \mathop{\arg\max}\limits_{\pi(\cdot |s_t)\in \Pi^{s}}\min\limits_{k \in {1.. K}}{Q_{\theta_k}}(s_t,a)-\alpha\log\pi(a|s_t)$, }
\STATE{where $a \sim\pi(\cdot |s_t)$}
\ELSE
\STATE{$\pi_b \leftarrow \overline{\pi}_{tar}$}
\ENDIF
\STATE {$a_t\sim \pi_b(\cdot|s_t)$}
\RETURN{$a_t$}
\end{algorithmic}
\end{algorithm}

As $\overline{\pi}_{tar}$ is synchronized with ${\pi}_{tar}$ immediately after each policy update, the output of $\overline{\pi}_{tar}$ is the same as ${\pi}_{tar}$, so we use $\overline{\pi}_{tar}$ as the current target policy in Algorithm \ref{alg:alg1}.
At the beginning of the learning process, the Q-value estimation may be inaccurate, and this behavior policy could be regarded as the optimistic exploration towards the actions with overestimated Q values. 
To limit the off-policyness of the behavior policy, $\epsilon$ in Algorithm \ref{alg:alg1} needs to be a small value. However, when $\epsilon$ approaches $0$ too much, the behavior policy cannot take advantage of the guidance policy. To balance exploration and exploitation, we set $\epsilon=0.2$ in the experiments.

\subsection{Integrating Optimization Transfer and Behavior Transfer}
\label{all}

\begin{algorithm*}[htbp]
   \caption{IOB}
   \label{alg:MetaCUP}
\begin{algorithmic}[1]
   \STATE {\bfseries Require:} Source policies $\{\pi_1,\pi_2,...,\pi_n\}$, hyper-parameters $\lambda_{\pi}, \lambda_{\alpha}, \tau, {\overline{\mathcal{H}}}$, {$\beta_s$}, $\epsilon$
   \STATE Initialize replay buffer $\mathcal{D}$
   \STATE Initialize $\pi_{tar}$ with parameter $\phi$, entropy weight $\alpha$, critic $Q_{\theta_k}$, target critic $Q_{{\overline{\theta}_k}} \leftarrow Q_{\theta_k},$ for $k \in \{1..K\}$
   \STATE{$\overline{\pi}_{tar} \leftarrow \pi_{tar}$, $\Pi^{s}\leftarrow\{\pi_1(\cdot |s),\pi_2(\cdot |s),...,\pi_n(\cdot |s),\overline{\pi}_{tar}(\cdot |s)\}$}
   \WHILE{not done} 
   \FOR{each environment step}
   \STATE{$a_t \leftarrow$ Behavior-$\pi(\epsilon, s_t, \Pi_s, \overline{\pi}_{tar}, \{Q_{\theta_k}|k \in{1..K}\})$}
   \STATE $s_{t+1} \sim p(s_{t+1}|s_t,a_t)$
   \STATE $\mathcal{D} \leftarrow \mathcal{D} \cup \{s_t,a_t,r(s_t,a_t),s_{t+1}\}$
   \STATE{$s_{t} \leftarrow s_{t+1}$}
   \ENDFOR
   \FOR{each gradient step}
   \STATE{Sample $K$ minibatches from $\mathcal{D}$, and update the critic networks independently}
   \STATE $\overline{\theta}_k \leftarrow \tau \theta_k + (1-\tau) \overline{\theta}_k$ for $i \in \{1..K\}$
   \STATE Sample minibatch $b$ from $\mathcal{D}$ to update $\pi_{tar}$ and $\alpha$
   \STATE {Query the  action probabilities $\{\pi_1(\cdot|s),\pi_2(\cdot|s),...,\pi_n(\cdot|s), \overline{\pi}_{tar}(\cdot |s)\}$ for state $s$ in $b$}
   \STATE {Compute expected advantages according to Eq. \eqref{ea}, form $\pi_g$ according to Eq. \eqref{arrgrgate}}
   \STATE $\phi \leftarrow \phi - \lambda_{\pi} \hat{\nabla}_{\phi} {L_{\pi}(\pi_{tar})}$
   \STATE $\alpha \leftarrow \alpha - \lambda_{\alpha}\hat{\nabla}_{\alpha}L_{entropy}(\alpha)$
   \STATE{Synchronize $\Pi_s$ with the updated $\pi_{tar}$}
   \ENDFOR
   \ENDWHILE
   \RETURN{$\pi_{tar}$}
\end{algorithmic}
\end{algorithm*}

The pseudo-code of IOB is presented in Algorithm \ref{alg:MetaCUP}. When interacting with the environment, the agent probabilistically utilizes  the guidance policy to collect samples (Line:6-11). During the policy updates, the guidance policy regularizes the direction of target policy optimization to achieve more efficient learning (Line:15-18).
After each policy update, $\overline{\pi}_{tar}$ in the policy set $\Pi_s$ is synchronized from the target policy $\pi_{tar}$ (Line 20).
Furthermore, as the guidance selection heavily depends on the learned critic, to select an effective guidance policy, we need a well-estimated Q function. Therefore, we employ the critic ensemble technique, and use different data to update multiple Q networks (Line:13-14).

\subsection{Theoretical Analysis}
\label{theory}
Note that we can hardly acquire the exact Q values to select the guidance policy during learning, and the Q values are estimated with function approximation in deep RL. In this subsection, we provide a theoretical analysis that even with an approximated Q function, we could form the guidance policy, and by regularizing the target policy to mimic the guidance policy, the target policy learning is guaranteed to achieve a monotonic improvement. 

\begin{thm}
Let {}$\widetilde{Q}_{\pi_{tar}}$ be an approximation of ${Q}_{\pi_{tar}}$, s.t., 
\begin{equation}
    |\widetilde{Q}_{\pi_{tar}}(s,a)-{Q}_{\pi_{tar}}(s,a)|\leq \mu \text{\ for\  all}\  s \in \mathcal{S}, a \in A.
\end{equation}
Define 
\begin{equation}
\begin{aligned}
   &\widetilde{\pi}_g(\cdot |s)=\mathop{\arg\max}\limits_{\pi(\cdot |s) \in \Pi^s}\mathbb{E}_{a \sim \pi(\cdot |s)}\left[\widetilde{Q}_{\pi_{tar}}(s,a)-\alpha\log\pi(a|s)\right], \\
  & \text{\ for\  all\ } s \in \mathcal{S}.
   \label{approximatepi}
   \end{aligned}
\end{equation}
Then, 
\begin{equation}
   V_{{\widetilde{\pi}_g}}(s)\geq V_{\pi_{tar}}(s) - \frac{2\mu}{1-\gamma} \text{\ for\  all\ } s \in \mathcal{S}.
\end{equation}
\label{thm2}
\end{thm}

Theorem 1 provides a way to obtain the guidance policy with approximated Q values, and the SAC method naturally learns that approximation, so that the guidance policy could be formed without training any additional components. In the following, we provide another theorem, which proves that policy improvement can be guaranteed if the target policy is optimized to stay close to the guidance policy. 

\begin{thm}
If 
\begin{equation}
   D_{KL}\left(\pi_{tar}^{l+1}(\cdot|s)||\widetilde{\pi}_g^l(\cdot|s)\right)\leq \delta\  for\  all\  s \in \mathcal{S},
\end{equation}
\ then 
\begin{equation}
\begin{aligned}
   V_{\pi_{tar}^{l+1}}(s) &\geq V_{\pi_{tar}^l}(s)-\frac{\sqrt{2\ln{2}\delta}(\widetilde{R}_{max}+\alpha\mathcal{H}_{max}^{l+1})}{(1-\gamma)^2}\\
 &  - \frac{2\mu+\alpha\widetilde{\mathcal{H}}_{max}}{1-\gamma}\  for\  all\  s \in \mathcal{S},
   \end{aligned}
\end{equation}
\label{thm3}
\end{thm}
where $\widetilde{\pi}_g^l$ is the guidance policy selected after the $l$-th policy update, $\pi_{tar}^l$ and $\pi_{tar}^{l+1}$ is the learned target policy after the $l$-th and $(l+1)$-th policy update. $\widetilde{R}_{max}=\max\limits_{s,a}|r(s,a)|$ is the largest possible absolute value of the reward, $\mathcal{H}_{max}^{l+1}=\max\limits_{s}\mathcal{H}(\pi_{tar}^{l+1}(\cdot|s))$ is the largest entropy of $\pi_{tar}^{l+1}$, and $\widetilde{\mathcal{H}}_{max}=\max\limits_{s}\left|\mathcal{H}(\pi_{tar}^{l}(\cdot|s))-\mathcal{H}(\pi_{tar}^{l+1}(\cdot|s))\right|$ is the largest possible absolute difference of the policy entropy.

\subsection{Transfer in a Continual RL Setting}
\label{crl}

In Continual RL\cite{khetarpal2020towards}, an agent will sequentially learn a series of tasks $\mathcal{Z}^{(1)},...,\mathcal{Z}^{(T_{max})}$, each corresponding to an individual MDP $\mathcal{Z}^{(t)}=(\mathcal{S}^{(t)},\mathcal{A}^{(t)},p^{(t)},r^{(t)},\gamma)$, while maintaining fixed constraints on computation and memory. The agent seeks out an optimal set of policy parameters $\{\phi^{(1)},...,\phi^{(T_{max})}\}$ to maximize the average rewards across all tasks. The Continual RL algorithm requires stability, i.e., the ability to prevent forgetting acquired skills, and plasticity, i.e., the ability to learn new skills quickly. Measures that enhance only one of these abilities often limit the other, resulting in a stability-plasticity dilemma.

To balance stability and plasticity as well as boost transfer, we combine the proposed method IOB with an advanced continual learning method, PackNet \cite{mallya2018packnet}. \cite{wolczyk2021continual} compared seven representative continual RL methods under the sequence of robotic arm tasks and showed that PackNet outperformed all other methods. PackNet develops a 
training-pruning-retraining procedure. After pruning, the parameters belonging to the previously learned policy are frozen, and only the  pruned parameters could be updated in the following tasks, so that the PackNet approach hardly forgets any policy.

However, the transfer ability of PackNet is limited, as it only considers parameter transfer by representing all the policies with one neural network. Integrating IOB with PackNet can potentially enhance the agent's ability to build upon prior knowledge. We wonder if IOB could improve the transfer performance of PackNet with optimization transfer and behavior transfer while maintaining the property of no forgetting. When applying IOB in the continual learning setting, we treat all the previously learned policies as source policies, i.e., when learning in the $i$-th task, there are $i-1$ source policies. The pseudocode of integrating IOB with PackNet is presented in Algorithm \ref{alg:alg3}.

\begin{algorithm}[H]
\caption{Continual RL with IOB}\label{alg:alg3}
\begin{algorithmic}[1]

\STATE{{\bfseries Require:} Continuous reinforcement learning task sequences }$\mathcal{Z}^{(1)},...,\mathcal{Z}^{(T_{max})}$, hyper-parameters $\lambda_{\pi}$, $\lambda_{\alpha}$, $\tau$, ${\overline{\mathcal{H}}}$, {$\beta_s$}, $\epsilon$
\STATE{Initialize source policies $\Pi^{s}=\{\}$}
\FOR{each task $\mathcal{Z}^{(t)}$}
\STATE{$\pi_t$ = IOB($\Pi^{s}$, hyper-parameters $\lambda_{\pi}, \lambda_{\alpha}, \tau, {\overline{\mathcal{H}}}$, {$\beta_s$}, $\epsilon$)}
\STATE{Sort parameters in each layer}
\STATE{Prune the smallest $\frac{n-t}{n-t+1}\%$ parameters per layer}
\STATE{Retrain network to maintain performance on $\mathcal{Z}^{(t)}$}
\STATE{Freeze parameters for $\mathcal{Z}^{(t)}$ and never update them}
\STATE{Append $\pi_t$ to $\Pi^{s}$}
\ENDFOR
\RETURN{$\Pi^{s}$}

\end{algorithmic}
\end{algorithm}

 \section{Experiments}\label{sec5}

\begin{figure}[hbtp]
\centering
\includegraphics[width=3.2in]{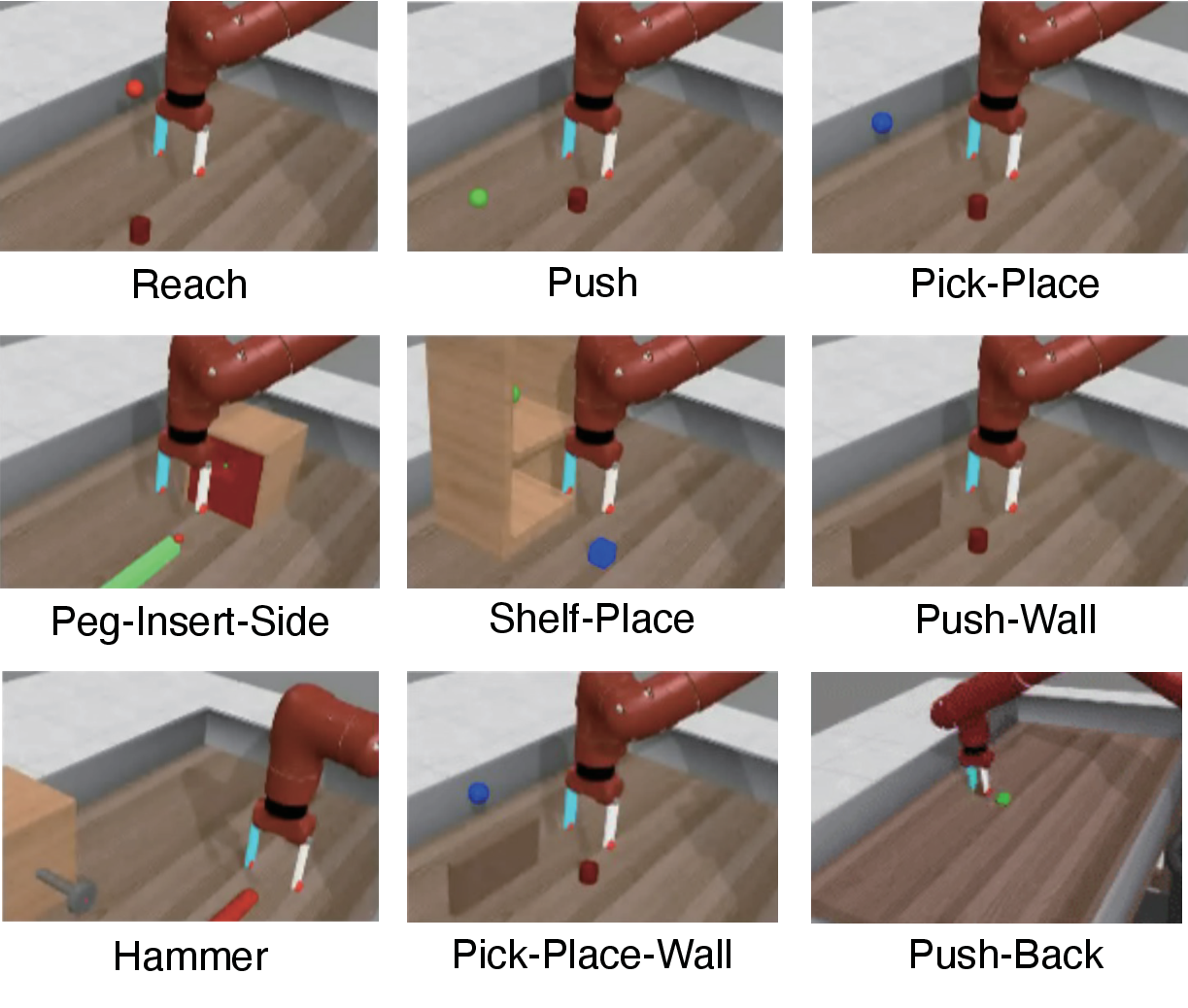}
\caption{Visualization of the tasks in the experiment section.}
\label{fig:env}
\end{figure}

We evaluate the proposed method IOB on Meta-World \cite{yu2020meta}, a popular RL benchmark composed of multiple robotic manipulation tasks. These tasks are both correlated (performed by the same Sawyer robot arm) and distinct (interacting with different objects and having different reward functions), and therefore serve as a proper evaluation benchmark for policy reuse. The source policies are obtained by training on three representative tasks: Reach, Push, and Pick-Place. We choose several complex tasks as target tasks, including Hammer, Peg-Insert-Side, Push-Wall, Pick-Place-Wall, Push-Back, and Shelf-Place. Among these target tasks, Hammer and Peg-Insert-Side require interacting with objects unseen in the source tasks. In Push-Wall and Pick-Place-Wall, there is a wall between the object and the goal. In Push-Back, the goal distribution is different from Push. In Shelf-Place, the robot is required to put a block on a shelf, and the shelf is unseen in the source tasks. Figure \ref{fig:env} visualizes these tasks and  the video demonstrations are available at \url{https://meta-world.github.io/}. Similar to the settings in \cite{yang2020multi}, in our experiments the goal position is randomly reset at the start of every episode, which increases the stochasticity in the environment and is thus more challenging. 

In this section, we first briefly introduce the baseline methods and the implementation details. Next, we show the comparison results in the Meta-World benchmark. Then, we analyze the guidance policy selection to dive into the reason why the proposed method could achieve positive transfer. After that, we perform several ablation studies to analyze the influence of the  components of IOB on transfer performance. Finally, we conduct a continual learning experiment, which demonstrates the effectiveness of the proposed method when the agent continuously learns in a sequence of tasks. 

\subsection{Baselines and Implementation Details}
We compare the proposed method with several representative transfer RL baseline algorithms and the backbone RL method in this work. 
\begin{itemize}
    \item CUP: A critic-guided policy reuse method without behavior transfer \cite{zhang2022cup}.
    \item HAAR: A hierarchical policy reuse method that simultaneously learns two-level policies  \cite{li2019hierarchical}.
    \item MAMBA: An optimization transfer method based on value function aggregation \cite{cheng2020policy}.
    \item MULTIPOLAR; A transfer RL method learning a weighted sum of the source policies' action probabilities and an additional network to predict residuals \cite{barekatain2021multipolar}.
    \item SAC: An off-policy actor-critic method with entropy maximization \cite{haarnoja2018soft2}.
\end{itemize}
\textbf{Implementation details:} 
To improve the computation efficiency of IOB, we store the outputs of the source policies in the replay buffer. When saving the buffer, we can query those source policies with batches of states. As those outputs are stored in the replay buffer, each state only needs to be queried one time, which leads to better computation efficiency.

Equation \eqref{arrgrgate} requires estimating the expectation over Q values. In practice, we obtain the estimation by sampling a few actions ($5$ actions) from the action distributions output by the source policies, and find that it is sufficient to accomplish a stable performance. 
The hyper-parameters used in the experiments are listed in Table \ref{tab:hyper}. The same set of hyper-parameters is used for all the tasks, and most hyper-parameters are adopted from \cite{sodhani2021multi}.

\begin{table}[htbp]
\caption{Hyper-parameter settings.}
  \label{tab:hyper}
\begin{tabular}{c|c}
\hline
    \multicolumn{1}{l}{Hyper-Parameter}  &\multicolumn{1}{l}{Value}    \\ \hline
\multicolumn{1}{l}{batch size }    & \multicolumn{1}{l}{ 1280 }\\
\multicolumn{1}{l}{non-linearity}        & \multicolumn{1}{l}{ReLU}     \\
\multicolumn{1}{l}{actor network structure}       & \multicolumn{1}{l}{3 fully-connected layers with 400 units} \\
\multicolumn{1}{l}{critic network structure}      & \multicolumn{1}{l}{3 fully-connected layers with 400 units} \\
\multicolumn{1}{l}{policy initialization}& \multicolumn{1}{l}{standard Gaussian}       \\
\multicolumn{1}{l}{learning rates for all networks}        & \multicolumn{1}{l}{3e-4}     \\
\multicolumn{1}{l}{optimizer}            & \multicolumn{1}{l}{Adam}     \\
\multicolumn{1}{l}{episode length (horizon)}           & \multicolumn{1}{l}{500}      \\
\multicolumn{1}{l}{discount}             & \multicolumn{1}{l}{0.99}     \\
\multicolumn{1}{l}{regularization rate $\beta_s$}          & \multicolumn{1}{l}{30} \\
\multicolumn{1}{l}{ensemble number $K$}          & \multicolumn{1}{l}{4}    \\ 
\multicolumn{1}{l}{guidance policy prob $\epsilon$}          & \multicolumn{1}{l}{0.2}    \\
\hline
\end{tabular}

\end{table}

\begin{figure*}[htbp]
\centering
\includegraphics[width=5.in]{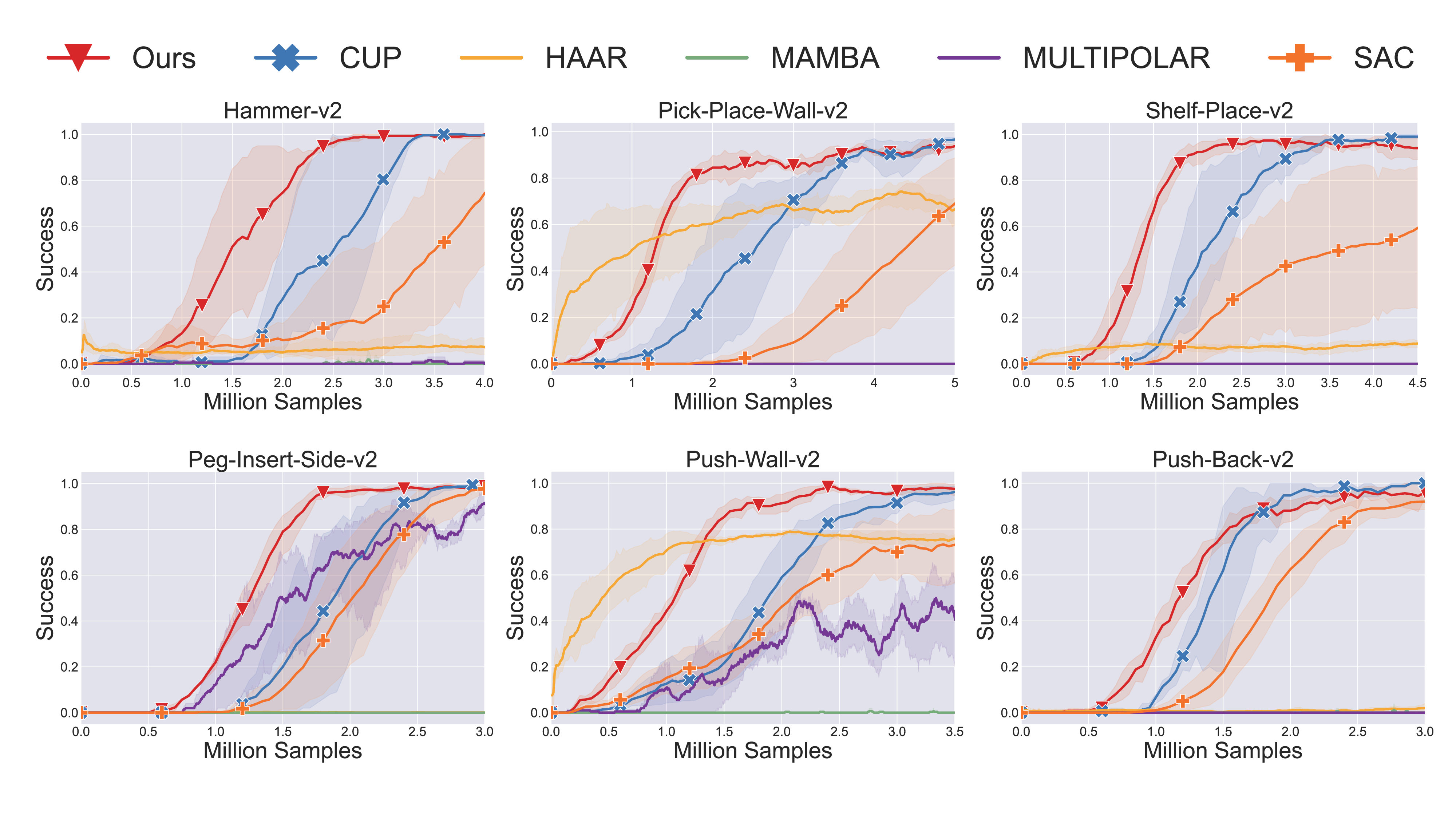}
\caption{Learning curves of the proposed method and the baselines on the tasks of the Meta-World benchmark. Videos comparing the policy learning processes of our method and SAC are at https://sites.google.com/view/iob-aamas.}
\label{curves}
\end{figure*}

\subsection{Comparison Results on Meta-World}

\begin{figure}[hbtp]
\centering
\includegraphics[width=3.2in]{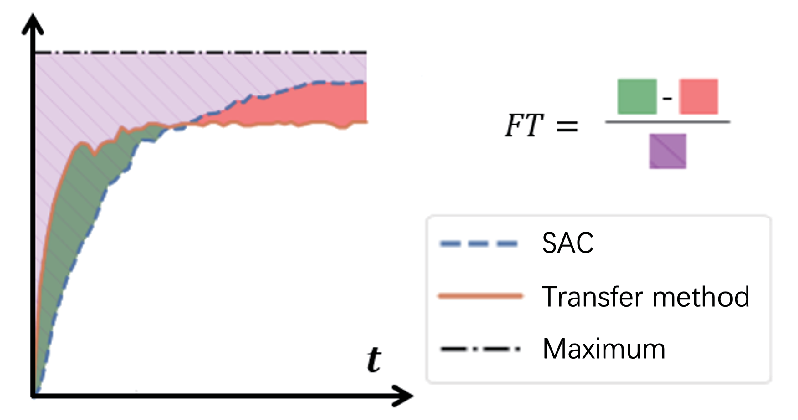}
\caption{An example of forward transfer $FT$\cite{wolczyk2021continual}.}
\label{fig:FT}
\end{figure}

We present the learning curves of all the methods in Figure \ref{curves}. Those learning curves are averaged over $5$ runs, and the success rates are averaged by $10$ evaluations. 
We also compare the \textit{forward transfer} of these methods in Table \ref{tab:FT}. Forward transfer, $FT$, is quantitative measure in transfer RL \cite{wolczyk2021continual}, which is defined as the normalized area between the learning curve of the transfer RL method and the learning curve of the reference method without transfer.
Specifically, we use SAC as the reference method. Figure \ref{fig:FT} is an example of forward transfer.

\begin{equation}
    FT = \frac{AUC_{trans} - AUC_{SAC}}{1-AUC_{SAC}}.
\end{equation}

\begin{table*}[htbp]
    \centering
    \caption{Forward Transfer Compared to SAC \label{tab:FT}}
    \resizebox{\linewidth}{!}{
        \begin{tabular}{c|ccccc}
            \hline
                                     & \textbf{Ours}  & \textbf{CUP} & \textbf{HAAR} & \textbf{MAMBA} & \textbf{MULTIPOLAR} \\ \hline
            \textbf{Hammer}          & {$\mathbf{0.51 \pm 0.09}$} & ${0.25 \pm 0.10}$  & ${-0.15 \pm 0.02}$ &  ${-0.23 \pm 0.0}$ & ${-0.23 \pm 0.0}$ \\
            \textbf{Peg-Insert-Side} & {$\mathbf{0.37 \pm 0.02}$} & ${0.08 \pm 0.13}$ & ${-0.45 \pm 0.0}$ & ${-0.45 \pm 0.0}$  & ${0.15 \pm 0.14}$  \\
            \textbf{Pick-Place-Wall} & {$\mathbf{0.61 \pm 0.01}$} &  ${0.38 \pm 0.12}$  & ${0.59 \pm 0.02}$ & ${-0.18  \pm 0.0}$ & ${-0.18 \pm 0.0}$  \\
            \textbf{Push-Wall}       & ${0.49 \pm 0.04}$  & ${0.13 \pm 0.12}$ & {$\mathbf{0.58 \pm 0.02}$}  &  ${-0.58  \pm 0.0}$ & ${-0.22 \pm 0.08}$ \\
            \textbf{Shelf-Place}     & {$\mathbf{0.56 \pm 0.03}$} & ${0.35 \pm 0.07}$ & ${-0.21 \pm 0.01}$  &  ${-0.31 \pm 0.0}$  & ${-0.31 \pm 0.0}$ \\
            \textbf{Push-Back}       & {$\mathbf{0.34 \pm 0.04}$} & ${0.26 \pm 0.07}$ &  ${-0.49 \pm 0.0}$ & ${-0.51 \pm 0.0}$ & $-0.51 \pm 0.0$ \\
            \hline
            \textbf{Avg}             & \textbf{0.48} & 0.24 & -0.02 & -0.38 & -0.22\\ \hline
            \end{tabular}
        }
\end{table*}

 As shown in Figure \ref{curves}, our approach has achieved significantly better performance than the baseline methods. 
 As lacking behavior transfer, the  CUP method is less efficient than the proposed approach. Note that the guidance policy selection for IOB and CUP is the same, comparing with CUP could be regarded as an ablation study for behavior transfer. 
 HAAR has a jump-start performance on Push-Wall and Pick-Place-Wall. However, due to the non-stationary issue induced by jointly training the high-level and low-level policies in the hierarchical structure, HAAR can hardly converge to a success rate close to $1$.   
 MULTIPOLAR achieves better performance on Push-Wall and Peg-Insert-Side than on the other tasks, because the Push source policy is useful on Push-Wall (implied by HAAR's good jump-start performance), and learning residuals on Peg-Insert-Side is easier (implied by SAC's fast learning). 
 In Pick-Place-Wall, the Pick-Place source policy is useful, but the residual is challenging to learn, so MULTIPOLAR does not work. For the remaining three tasks, the source policies are less useful, and therefore MULTIPOLAR fails on these tasks.
 The MAMBA method could hardly learn any successful policy, since accurately estimating the value functions for
 all source policies is not sample efficient.  

 Table \ref{tab:FT} demonstrates the forward transfer values for all the methods. In almost all the tasks, the proposed approach has achieved the largest forward transfer, which indicates that our approach generally works.  In the Push-Wall task, the forward transfer of HAAR is slightly larger than ours, but this slight outperformance is at the expense of a smaller convergent success rate, which is unfavorable.

\subsection{Analysis of Guidance Policy Selection}

\begin{figure}[hbtp]
\centering
\includegraphics[width=3.2in]{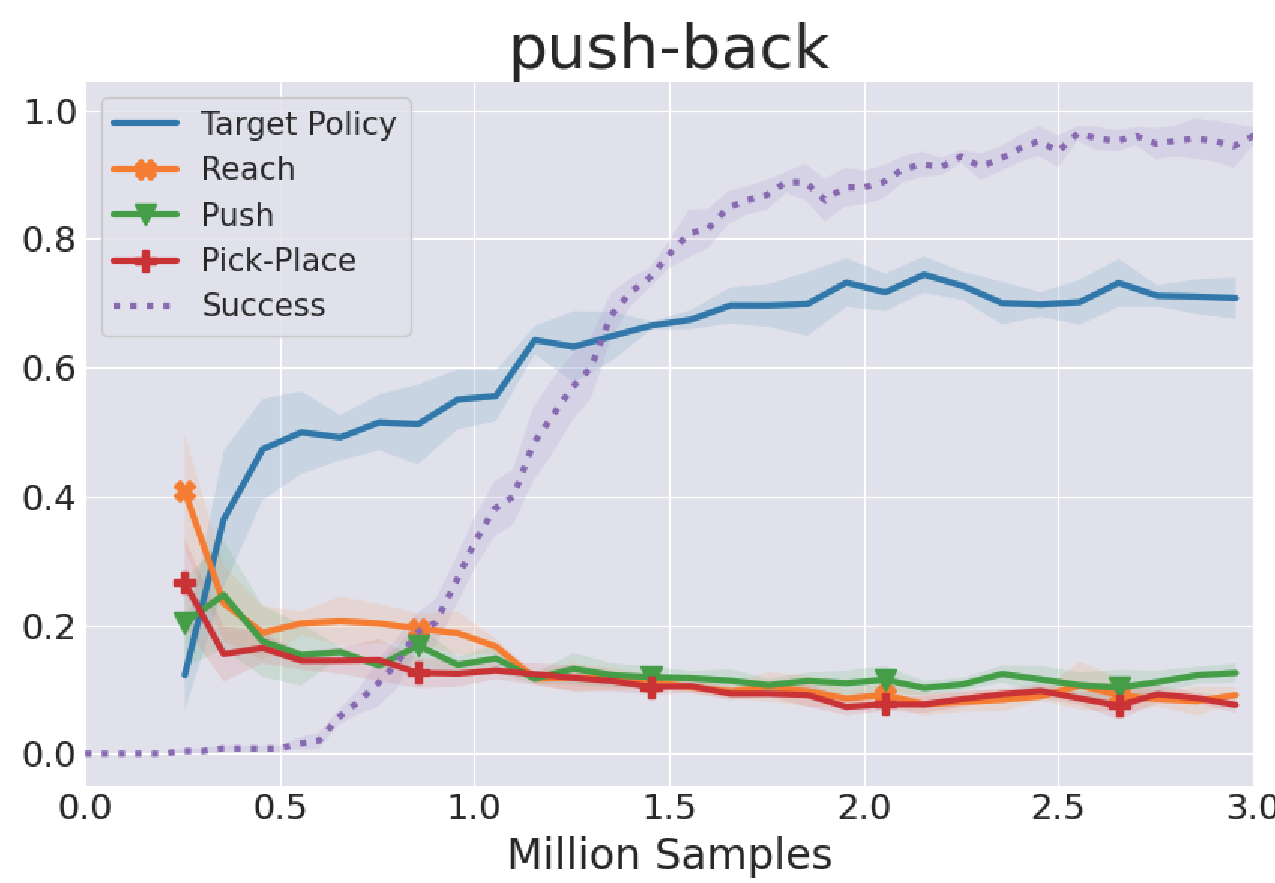}
\caption{Percentages of the source policies and the current target policy selected as the guidance policy during training in the Push-Back task. The dashed line is the success rate of the target policy in this task.}
\label{fig:weight}
\end{figure}

This subsection visualizes the guidance policy selection during the learning process of the proposed approach. Figure \ref{fig:weight} shows the percentages of the given source policies and the learned target policy being selected as the guidance policy throughout the training on the Push-Back task. Note that the critic network is randomly initialized, and after several optimization iterations, the critic could identify which action is more beneficial. Therefore,   after $0.25$M steps, we utilize the guidance policy selected with the critic to conduct transfer.

At the early stages of training, the source policies are selected more frequently as they have positive expected advantages. This indicates that they can be used to improve the current target policy. As the training proceeds and the target policy becomes better, the source policies are selected less frequently. Among these three source policies, Reach is chosen more frequently than the other two source policies, as in most manipulation tasks, the robot needs to first reach the target object. Figure \ref{fig:weight} validates that the proposed method could transfer knowledge from multiple source policies to facilitate the target task learning as well. 

\subsection{Ablation Study}
In this subsection, we conduct three ablation studies to answer the following questions: (1) Could the proposed method benefit from a larger source policy set, which includes more related source policies? (2) What is the influence of random source policies on the transfer performance of the proposed method? (3) Could behavioral transfer enhance transfer performance, and what is the proper value of $\epsilon$ in the evaluated tasks?

\subsubsection{Larger Source Policy Sets}

\begin{figure}[hbtp]
\centering
\includegraphics[width=3.2in]{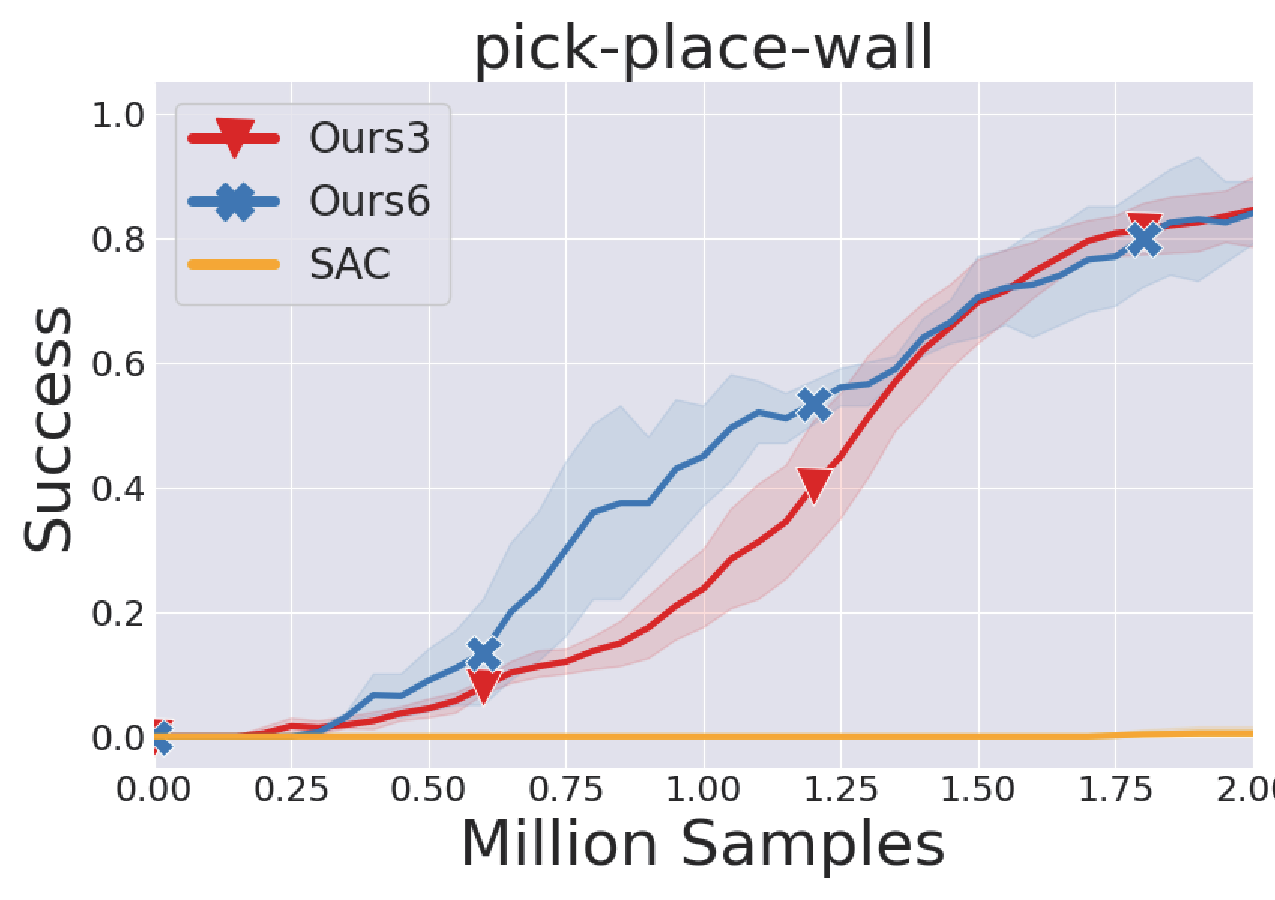}
\caption{Ablation study with different numbers of source policies on the Pick-Place-Wall task.}
\label{fig:6source}
\end{figure}

We evaluate the proposed method with a larger source policy set on the Pick-Place-Wall task. The original source policy set is expanded with three additional policies, which solve the Drawer-Close, Push-Wall and Coffee-Button tasks, i.e., the new source policy set is composed of $6$ policies. Figure \ref{fig:6source} provides the comparison results of our method with 3 source policies and with 6 source policies. The results show that our method is able to utilize the additional source policies to further improve the transfer performance. As the original policy set with three policies has already contained one useful source, Pick-Place, the improvement caused by the additional source policies is relatively slight.  

\subsubsection{Random Source Policies}

To investigate the robustness of the proposed method to the useless random source policies, we design a transfer setting with two source policy sets. One source policy contains three random policies, and the other contains the Reach policy and  three random policies.  As shown in Figure \ref{fig:random}, when no source policy is useful, the proposed method performs similarly to the SAC method, and its sample efficiency is almost unaffected by those random source policies. When there is one useful source policy, the proposed method could efficiently utilize it to improve the learning performance, even if a lot of useless random source policies exist. 

\begin{figure}[hbtp]
\centering
\includegraphics[width=3.2in]{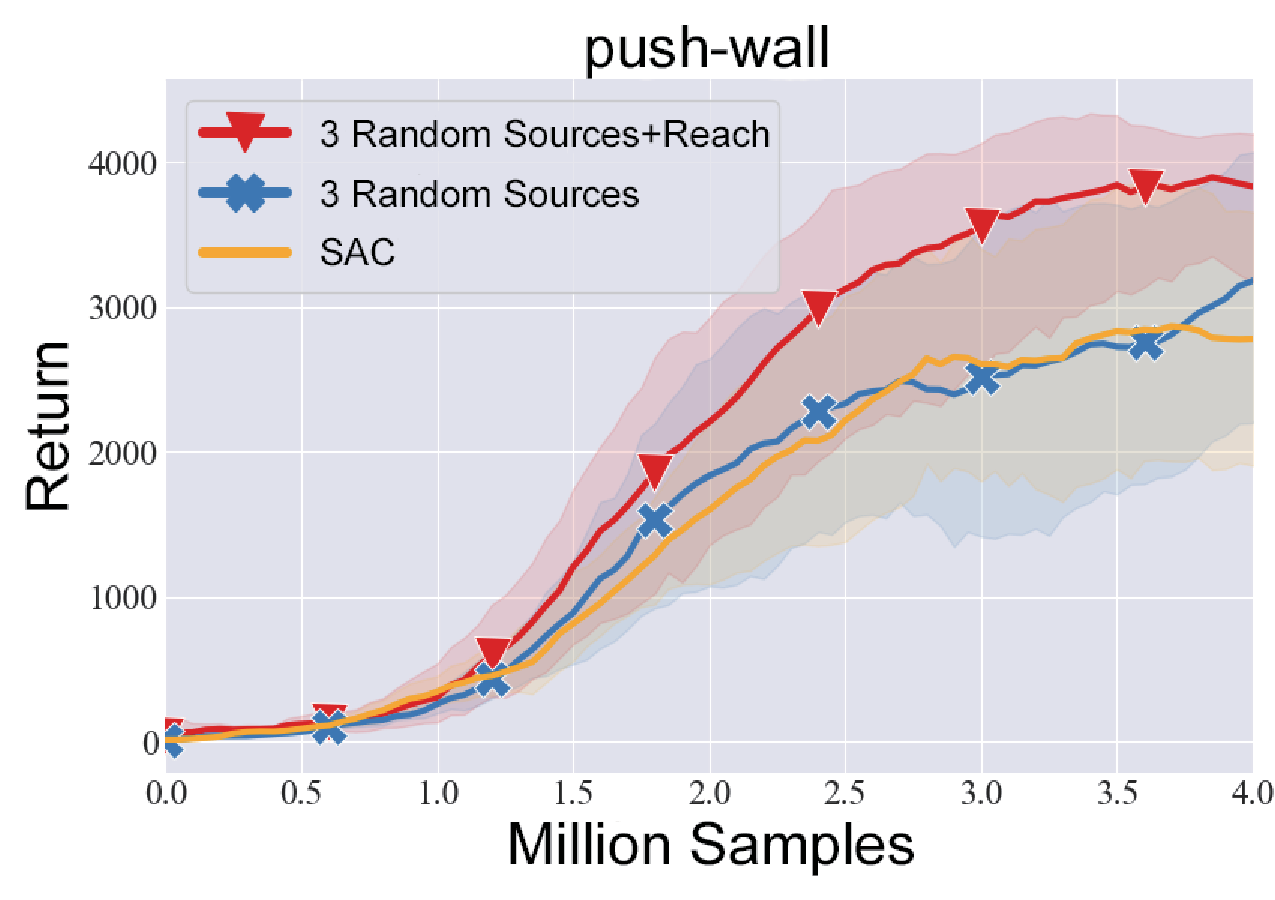}
\caption{Ablation studies on IOB’s sensitivity to useless source policies.}
\label{fig:random}
\end{figure}

\subsubsection{Selection of Hyperparameter $\epsilon$ }

In order to demonstrate the effectiveness of the behavior transfer, we evaluate the transfer performance of IOB under different $\epsilon$ in the pick-place-wall-v2 task. We have visualized the early training process of this task in Figure 8. Evidently, without behavior transfer, the agent cannot obtain high-value samples solely through self-exploration. Conversely, appropriate behavior transfer can effectively enhance transfer performance; however, excessive behavioral transfer can result in performance decline. This can be attributed to the model learned almost exclusively under offline settings, making it more challenging to evaluate the Q function accurately. Notably, accurate evaluation of the Q-function is crucial for the IOB method.

\begin{figure}[hbtp]
\centering
\includegraphics[width=3.2in]{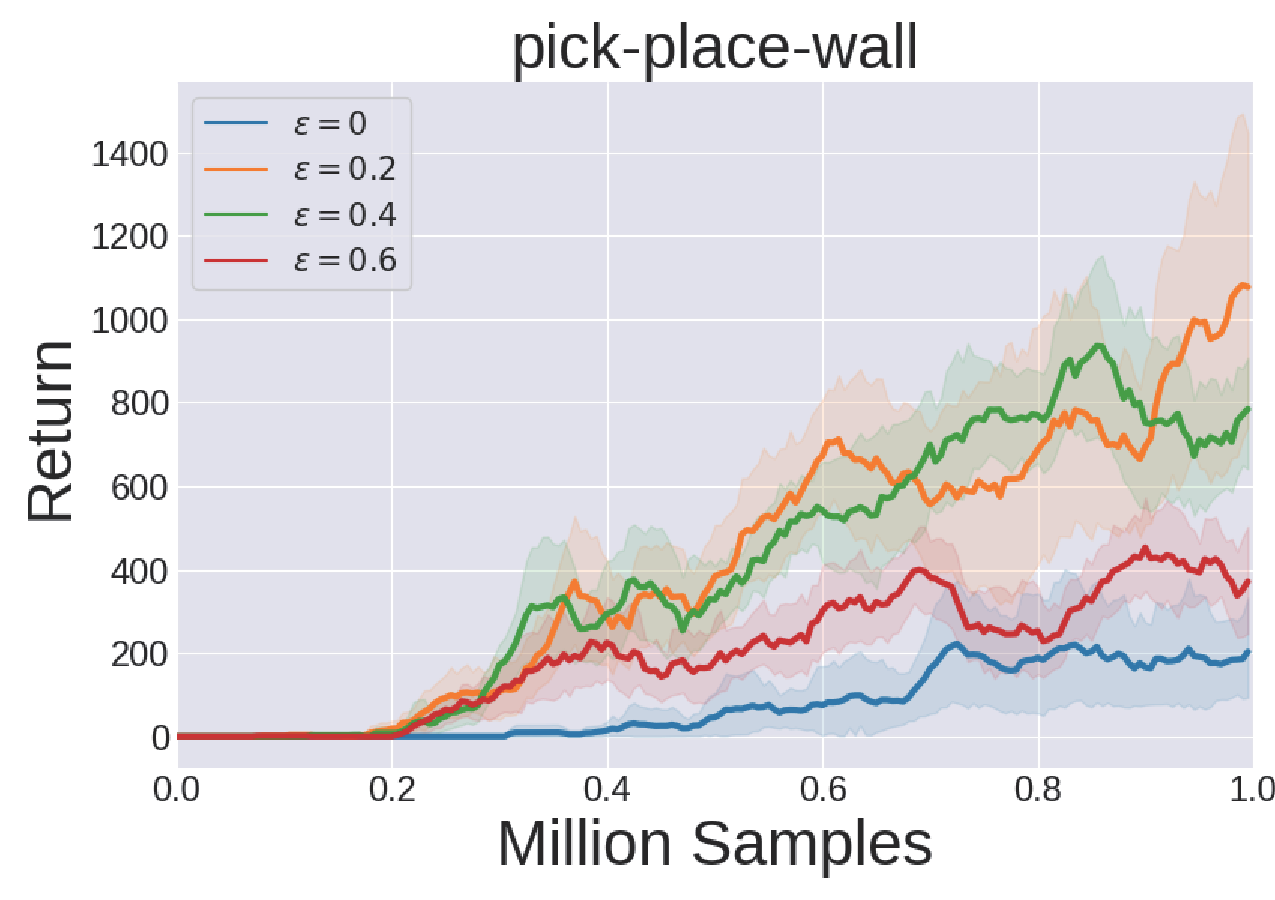}
\caption{Ablation study on behavioral transfer effects over a range of $\epsilon$ magnitudes.}
\label{fig:random}
\end{figure}

\subsection{Transfer in a Continual Learning Setting}
\label{continual}
\begin{figure*}[htbp]
\centering
\includegraphics[width=5.0in]{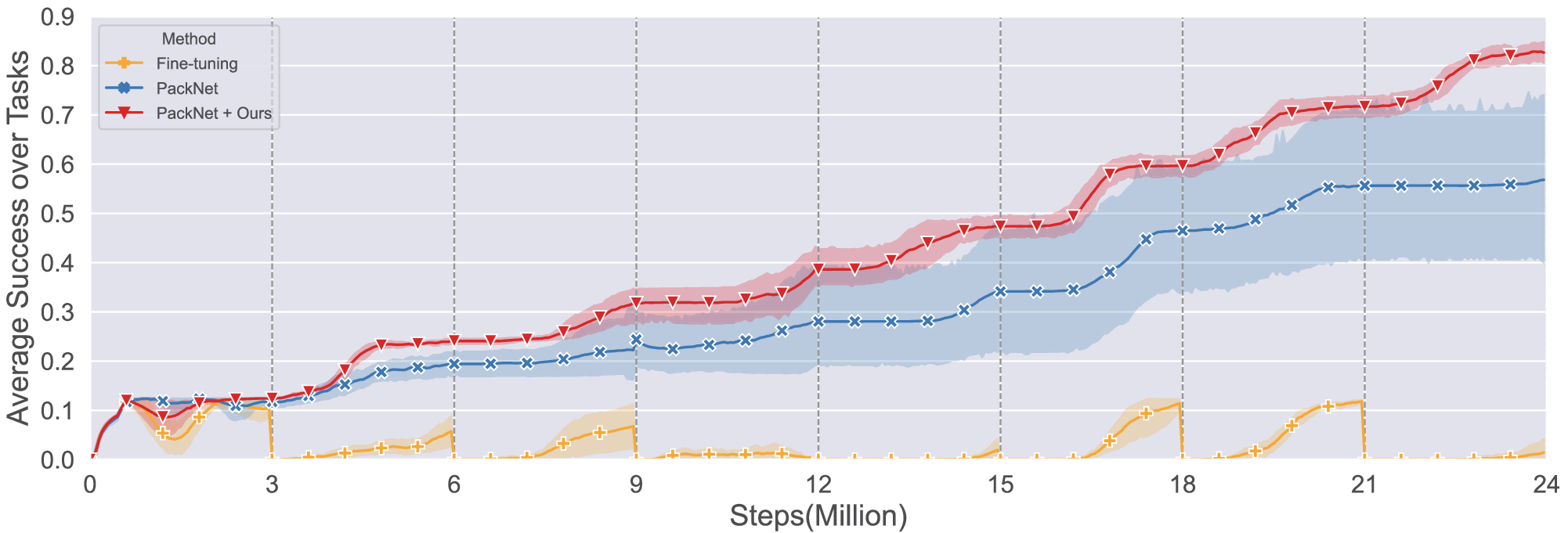}
\caption{Average success rates over the $8$ tasks of Fine-tuning, PackNet, and applying the proposed method IOB to PackNet.}
\label{fig:cl}
\end{figure*}

To validate whether the proposed method could enhance transfer in a continual learning setting, we conduct a continual RL experiment. In this experiment, an agent sequentially learns policies in eight tasks: Reach-v2, Push-v2, Pick-Place-v2, Hammer-v2, Shelf-Place-v2, Peg-Insert-Side-v2, Push-Wall-v2, Pick-Place-Wall-v2. In each task, the sample budget is 3 million samples. We compare our method with PackNet and a naive continual learning baseline, i.e., fine-tuning the previously learned policy.
As continual learning requires the agent to have the ability of both learning new policies and not forgetting previously learned policies, we evaluate these policies on all the eight tasks, and report the average success rates over tasks in Figure  \ref{fig:cl}. 

The results in Figure \ref{fig:cl} demonstrate that the PackNet method remember all the previously learned policy, as the average success rate curve of PackNet has not dropped in the learning process. Compared with PackNet, the forgetting phenomenon of fine-tuning is very obvious. At the beginning of learning in a new task (the dashed lines), the average success rate of fine-tuning drops immediately, which indicates that the fine-tuning method forgets the previous policy. Taking advantage of the parameter isolation technique of PackNet, our method could remember all the previously learned policies as well. Beyond that, our method achieves faster learning and larger convergence success rates via behavior transfer and optimization transfer. After training in the $8$ tasks with $24$ ($3\times 8$) million steps, our method has achieved an average success rate of more than $0.8$. 

In Table \ref{tab:cl}, we provide the average convergent success rate at $T$ steps ($T=24$ million), average forward transfer, and average forgetting over the $8$ tasks. The forgetting of the $i$-th policy is measured by the drop of the success rate from the end of training in the $i$-th task to the end of the total learning process, 
\begin{equation}
    Forgetting_i = SR(i\cdot\Delta) - SR(T),
\end{equation}
where $SR$ denotes the evaluation success rate, $\Delta$ is the sample budget for each task ($3$ million). 
From Table \ref{tab:cl}, we can see that our method has achieved the largest forward transfer, the least forgetting and the best convergent success rate among the three methods. 
However, compared to the average forward transfer over the tasks in Table \ref{tab:FT}, we find that the forward transfer of combining our method with PackNet is smaller, which indicates that the effect of reusing parameters may be negative when conducting optimization transfer and behavior transfer. 
 The PackNet method demonstrates no forgetting, but its forward transfer is smaller than its variant combined with our method. 

 \begin{table}[htbp]
    \centering
    \caption{Continual Learning Results \label{tab:cl}}
        \begin{tabular}{cccc}
            \hline
    & \textbf{Success}  & \textbf{Forgetting} & \textbf{Transfer}  \\ \hline
\textbf{Fine-tuning}  & {${0.02 \pm 0.02}$} & ${0.55 \pm 0.12}$  & ${-0.06 \pm 0.14}$  \\
\textbf{PackNet} & {${0.57 \pm 0.20}$} & $\mathbf{0.00 \pm 0.0}$ & ${0.07 \pm 0.14}$ \\
\textbf{PackNet+Ours} & {$\mathbf{0.83 \pm 0.02}$} &  $\mathbf{0.01 \pm 0.01}$  & $\mathbf{0.24 \pm 0.13}$  \\ \hline
            \end{tabular}
\end{table}



\section{Conclusion}\label{sec7}

To transfer knowledge efficiently from multiple source policies to a related target task, we propose a novel transfer RL method that conducts guidance policy selection without training any extra component. By utilizing the Q function as a natural evaluation of the source policies, the proposed method selects the policy with the largest one-step improvement over the current target policy as the guidance policy. The selected guidance policy is used to regularize the policy optimization process to accomplish optimization transfer. Meanwhile, to boost transfer performance further with behavior transfer, we combine the guidance policy with the learned policy as the behavior policy. Benefiting from the effective guidance policy selection and the integration of optimization transfer and  behavior transfer, the proposed method significantly outperforms the state-of-the-art transfer RL baselines on the benchmark tasks. Beyond that, we provide a theoretical analysis that with the proposed policy optimization technique, the target policy is guaranteed to be improved monotonically.

As for future work, we consider relaxing the assumption in this paper that all the source policies and the target policy share the same state and action spaces. This assumption somehow limits the applications of the proposed method to more general circumstances. Aligning different state and action spaces is a challenging problem. Previous research works investigate this problem with a high-level structure \cite{wan2020mutual,zhang2020learning,heng2022cross,van2020mdp,van2020plannable}, and some inspiration could be taken from those previous works. 

\backmatter

\bmhead{Acknowledgments}

This work is supported by the Key Program of the National Natural Science Foundation of China (Grant No.51935005), Basic Research Project (Grant No.JCKY20200603C010), Natural Science Foundation of Heilongjiang Province of China (Grant No.LH2021F023), as well as Science and Technology Planning Project of Heilongjiang Province of China (Grant No.GA21C031), and China Academy of Launch Vehicle Technology (CALT2022-18).

\begin{appendices}


\section*{Proof of Theorem 1}
\textit{Proof.} As $|\widetilde{Q}_{\pi_{tar}}(s,a)-{Q}_{\pi_{tar}}(s,a)|\leq \mu \text{\ for\  all}\  s \in \mathcal{S}, a \in A$, we have that for all $s \in \mathcal{S}$, the difference between the true value function $V_{\pi_{tar}}$ and the approximated value function $\widetilde{V}_{\pi_{tar}}$ is bounded:
\begin{align}
    & V_{\pi_{tar}}(s) \notag  \\
    &=\mathbb{E}_{a \sim \pi_{tar}(\cdot|s)}\left[Q_{\pi_{tar}}(s,a)-\alpha\log\pi_{tar}(a|s)\right] \notag \\
    & \leq \mathbb{E}_{a \sim \pi_{tar}(\cdot|s)}\left[\widetilde{Q}_{\pi_{tar}}(s,a)-\alpha\log\pi_{tar}(a|s)+\mu\right]  \notag \\
    & = \widetilde{V}_{\pi_{tar}}(s)+ \mu. \notag
\end{align}
\ As $\pi_{tar}(\cdot|s)$ is contained in $\Pi^s$, with $\widetilde{\pi}_{g}$ defined in Eq. \eqref{approximatepi}, it is obvious that for all $s \in \mathcal{S}$, 
\begin{equation}
\begin{aligned}
&\mathbb{E}_{a \sim \widetilde{\pi}_{g}} (\cdot|s)\left[\widetilde{Q}_{\pi_{tar}}(s,a)-\alpha\log\widetilde{\pi}_g(a|s)\right] \geq \\
&\mathbb{E}_{a \sim \pi_{tar} (\cdot|s)}\left[\widetilde{Q}_{\pi_{tar}}(s,a)-\alpha\log\pi_{tar}(a|s)\right]=\widetilde{V}_{\pi_{tar}}(s).
\end{aligned}
\end{equation}
Then for all $s_i \in \mathcal{S}$,
\begin{align}
    & V_{\pi_{tar}}(s_i)  \leq \widetilde{V}_{\pi_{tar}}(s_i)+ \mu \notag \\
    & \leq \mathbb{E}_{a_i \sim \widetilde{\pi}_{g}(\cdot|s_i)}[\widetilde{Q}_{\pi_{tar}}(s_i,a_i)-\alpha\log\widetilde{\pi}_{g}(a_i|s_i)] + \mu  \notag \\
    & \leq \mathbb{E}_{a_i \sim \widetilde{\pi}_{g}}(\cdot|s_i)[{Q}_{\pi_{tar}}(s_i,a_i)-\alpha\log\widetilde{\pi}_{g}(a_i|s_i)]+ 2\mu  \notag \\
        &=\mathbb{E}_{a_i \sim \widetilde{\pi}_{g}}(a|s_i)[r(s_i,a_i)-\alpha\log\widetilde{\pi}_{g}(a_i|s_i) +\gamma V_{\pi_{tar}}(s_{i+1})] \notag \\
   &  \ \ \ + 2\mu \notag \\
     & \vdots  \notag \\
     & \leq \mathbb{E}_{\widetilde{\pi}_{g}}[\sum_{\tau=0}^{\infty}\gamma^{\tau}(r(s_{i+\tau},a_{i+\tau})-\alpha\log\widetilde{\pi}_{g}(a_{i+\tau}|s_{i+\tau}))] \notag \\
     & \ \ \ + 2 \sum_{\tau=0}^{\infty}\gamma^{\tau}\mu \notag \\
     & =V_{\widetilde{\pi}_{g}} (s_i)+\frac{2\mu}{1-\gamma}. \notag
\end{align}

\section*{Proof of Theorem 2}
\textit{Proof.} According to the Pinsker’s inequality \cite{fedotov2003refinements}, $D_{KL}(\pi_{tar}^{l+1}(\cdot|s)||\widetilde{\pi}_{g}^l(\cdot|s))\geq \frac{1}{2\ln2}||\pi_{tar}^{l+1}(\cdot|s)-\widetilde{\pi}_{g}^l(\cdot|s)||_1^2$, where $||\cdot||_1$ is the L1 norm. So we have that for all $s \in \mathcal{S}$, $||\pi_{tar}^{l+1}(\cdot|s)-\widetilde{\pi}_{g}^l(\cdot|s)||_1 \leq \sqrt{2\ln2 \delta}$. According to the Performance Difference Lemma \cite{kakade2002approximately}, we have that for all $s \in \mathcal{S}$:
\begin{align}
    &V_{\widetilde{\pi}_{g}^l}(s)-V_{\pi_{tar}^{l+1}}(s) \notag  \\
    & = \frac{1}{1-\gamma}\mathbb{E}_{s' \sim \rho_{s}^{\widetilde{\pi}_{g}^l}}(s')  [\mathbb{E}_{a \sim \widetilde{\pi}_{g}^l(\cdot|s')}[Q_{\pi_{tar}^{l+1}}(s',a)-\alpha\log\widetilde{\pi}_{g}^l(a|s)] \notag \\
    & \ \ \ -\mathbb{E}_{a \sim \widetilde{\pi}_{tar}^{l+1}(\cdot|s')}[Q_{\pi_{tar}^{l+1}}(s',a)-\alpha\log\widetilde{\pi}_{tar}^{l+1}(a|s)]] \notag \\
    & \leq \frac{1}{1-\gamma}\max\limits_{s' \in \mathcal{S}}[\mathbb{E}_{a \sim \widetilde{\pi}_{g}^l}(\cdot|s')[Q_{\pi_{tar}^{l+1}}(s',a)] \notag \\
   & \ \ \ -\mathbb{E}_{a \sim \pi_{tar}^{l+1}(\cdot|s')}[Q_{\pi_{tar}^{l+1}}(s',a)]] \notag \\
   & \ \ \ +\frac{\alpha}{1-\gamma}\max\limits_{s'' \in \mathcal{S}}\left|\mathcal{H}(\widetilde{\pi}_{g}^l(\cdot|s''))-\mathcal{H}(\pi_{tar}^l(\cdot|s''))\right| \notag \\
    & = \frac{1}{1-\gamma}\max\limits_{s' \in \mathcal{S}}\int\left(\widetilde{\pi}_{g}^l(\cdot|s)-\pi_{tar}^{l+1}(a|s)\right)Q_{\pi_{tar}^{l+1}}(s',a)da \notag \\
    & \ \ \ +\frac{\alpha}{1-\gamma}\widetilde{\mathcal{H}}_{max} \notag \\
    & \leq\frac{1}{1-\gamma} \max\limits_{s' \in \mathcal{S}}\int\left|\widetilde{\pi}_{g}^l(a|s)-\pi_{tar}^{l+1}(a|s)\right|\cdot\left|Q_{\pi_{tar}^{l+1}}(s',a)\right|da \notag \\
   & \ \ \ +\frac{\alpha}{1-\gamma}\widetilde{\mathcal{H}}_{max} \notag \\
    &\leq \frac{1}{1-\gamma} \max\limits_{s' \in \mathcal{S}}\int\left|\widetilde{\pi}_{g}^l(a|s)-\pi_{tar}^{l+1}(a|s)\right|\cdot\frac{\widetilde{R}_{max}+\alpha\mathcal{H}_{max}^{l+1}}{1-\gamma}da \notag \\
    & \ \ \ +\frac{\alpha}{1-\gamma}\widetilde{\mathcal{H}}_{max} \notag 
     \end{align}
\begin{align}
    & = \frac{\widetilde{R}_{max}+\alpha\mathcal{H}_{max}^{l+1}}{(1-\gamma)^2}\max\limits_{s' \in \mathcal{S}}||\widetilde{\pi}_{g}^l(\cdot|s)-\pi_{tar}^{l+1}(\cdot|s)||_1 \notag \\
   & \ \ \  +\frac{\alpha}{1-\gamma}\widetilde{\mathcal{H}}_{max} \notag \\
    & \leq \frac{\sqrt{2\ln2\delta}(\widetilde{R}_{max}+\alpha\mathcal{H}_{max}^{l+1})+\alpha(1-\gamma)\widetilde{\mathcal{H}}_{max}}{(1-\gamma)^2}, \notag \\
\end{align}
where $\rho_{s}^{\widetilde{\pi}_{g}^l}(s')=(1-\gamma)\sum_{t=0}^{\infty}\gamma^tp(s_t=s'|s_0=s,\widetilde{\pi}_{g}^l)$ is the normalized discounted state occupancy distribution. Note that
\begin{align}
&|Q_{\pi_{tar}^{l+1}}(s,a)|\notag \\
&=|\mathbb{E}_{\pi_{tar}^{l+1}}[\sum_{i=0}^{\infty}\gamma^i(r(s_{\tau+i},a_{\tau+i}) \notag \\
& \ \ \ -\alpha\log\pi_{tar}^{l+1}(\cdot|s))|s_\tau=s,a_\tau=a]| \notag \\
&\leq\mathbb{E}_{\pi}[\sum_{i=0}^{\infty}\gamma^i(\widetilde{R}_{max}+\gamma\mathcal{H}_{max}^{l+1})] \\
&=\frac{\widetilde{R}_{max}+\alpha\mathcal{H}_{max}^{l+1}}{1-\gamma}.
\end{align}
\ Eventually, we have 
\begin{align}
&V_{\pi_{tar}^{l+1}}(s) \notag \\
&\geq V_{\widetilde{\pi}_{g}^l}(s) - \frac{\sqrt{2\ln2\delta}(\widetilde{R}_{max}+\alpha\mathcal{H}_{max}^{l+1})+\alpha(1-\gamma)\widetilde{\mathcal{H}}_{max}}{(1-\gamma)^2} \notag \\
&\geq V_{\pi_{tar}^l}(s)-\frac{\sqrt{2\ln2\delta}(\widetilde{R}_{max}+\alpha\mathcal{H}_{max}^{l+1})}{(1-\gamma)^2}- \frac{2\mu+\alpha\widetilde{\mathcal{H}}_{max}}{1-\gamma}.
\end{align}




\end{appendices}


\bibliography{sn-bibliography}


\begin{thebibliography}{48}
\ifx \bisbn   \undefined \def \bisbn  #1{ISBN #1}\fi
\ifx \binits  \undefined \def \binits#1{#1}\fi
\ifx \bauthor  \undefined \def \bauthor#1{#1}\fi
\ifx \batitle  \undefined \def \batitle#1{#1}\fi
\ifx \bjtitle  \undefined \def \bjtitle#1{#1}\fi
\ifx \bvolume  \undefined \def \bvolume#1{\textbf{#1}}\fi
\ifx \byear  \undefined \def \byear#1{#1}\fi
\ifx \bissue  \undefined \def \bissue#1{#1}\fi
\ifx \bfpage  \undefined \def \bfpage#1{#1}\fi
\ifx \blpage  \undefined \def \blpage #1{#1}\fi
\ifx \burl  \undefined \def \burl#1{\textsf{#1}}\fi
\ifx \doiurl  \undefined \def \doiurl#1{\url{https://doi.org/#1}}\fi
\ifx \betal  \undefined \def \betal{\textit{et al.}}\fi
\ifx \binstitute  \undefined \def \binstitute#1{#1}\fi
\ifx \binstitutionaled  \undefined \def \binstitutionaled#1{#1}\fi
\ifx \bctitle  \undefined \def \bctitle#1{#1}\fi
\ifx \beditor  \undefined \def \beditor#1{#1}\fi
\ifx \bpublisher  \undefined \def \bpublisher#1{#1}\fi
\ifx \bbtitle  \undefined \def \bbtitle#1{#1}\fi
\ifx \bedition  \undefined \def \bedition#1{#1}\fi
\ifx \bseriesno  \undefined \def \bseriesno#1{#1}\fi
\ifx \blocation  \undefined \def \blocation#1{#1}\fi
\ifx \bsertitle  \undefined \def \bsertitle#1{#1}\fi
\ifx \bsnm \undefined \def \bsnm#1{#1}\fi
\ifx \bsuffix \undefined \def \bsuffix#1{#1}\fi
\ifx \bparticle \undefined \def \bparticle#1{#1}\fi
\ifx \barticle \undefined \def \barticle#1{#1}\fi
\bibcommenthead
\ifx \bconfdate \undefined \def \bconfdate #1{#1}\fi
\ifx \botherref \undefined \def \botherref #1{#1}\fi
\ifx \url \undefined \def \url#1{\textsf{#1}}\fi
\ifx \bchapter \undefined \def \bchapter#1{#1}\fi
\ifx \bbook \undefined \def \bbook#1{#1}\fi
\ifx \bcomment \undefined \def \bcomment#1{#1}\fi
\ifx \oauthor \undefined \def \oauthor#1{#1}\fi
\ifx \citeauthoryear \undefined \def \citeauthoryear#1{#1}\fi
\ifx \endbibitem  \undefined \def \endbibitem {}\fi
\ifx \bconflocation  \undefined \def \bconflocation#1{#1}\fi
\ifx \arxivurl  \undefined \def \arxivurl#1{\textsf{#1}}\fi
\csname PreBibitemsHook\endcsname

\bibitem[\protect\citeauthoryear{Guberman and
  Greenfield}{1991}]{guberman1991learning}
\begin{barticle}
\bauthor{\bsnm{Guberman}, \binits{S.R.}},
\bauthor{\bsnm{Greenfield}, \binits{P.M.}}:
\batitle{Learning and transfer in everyday cognition}.
\bjtitle{Cognitive Development}
\bvolume{6}(\bissue{3}),
\bfpage{233}--\blpage{260}
(\byear{1991})
\end{barticle}
\endbibitem

\bibitem[\protect\citeauthoryear{Silver et~al.}{2017}]{silver2017mastering}
\begin{barticle}
\bauthor{\bsnm{Silver}, \binits{D.}},
\bauthor{\bsnm{Schrittwieser}, \binits{J.}},
\bauthor{\bsnm{Simonyan}, \binits{K.}},
\bauthor{\bsnm{Antonoglou}, \binits{I.}},
\bauthor{\bsnm{Huang}, \binits{A.}},
\bauthor{\bsnm{Guez}, \binits{A.}},
\bauthor{\bsnm{Hubert}, \binits{T.}},
\bauthor{\bsnm{Baker}, \binits{L.}},
\bauthor{\bsnm{Lai}, \binits{M.}},
\bauthor{\bsnm{Bolton}, \binits{A.}}, \betal:
\batitle{Mastering the game of go without human knowledge}.
\bjtitle{nature}
\bvolume{550}(\bissue{7676}),
\bfpage{354}--\blpage{359}
(\byear{2017})
\end{barticle}
\endbibitem

\bibitem[\protect\citeauthoryear{Vinyals et~al.}{2019}]{vinyals2019grandmaster}
\begin{barticle}
\bauthor{\bsnm{Vinyals}, \binits{O.}},
\bauthor{\bsnm{Babuschkin}, \binits{I.}},
\bauthor{\bsnm{Czarnecki}, \binits{W.M.}},
\bauthor{\bsnm{Mathieu}, \binits{M.}},
\bauthor{\bsnm{Dudzik}, \binits{A.}},
\bauthor{\bsnm{Chung}, \binits{J.}},
\bauthor{\bsnm{Choi}, \binits{D.H.}},
\bauthor{\bsnm{Powell}, \binits{R.}},
\bauthor{\bsnm{Ewalds}, \binits{T.}},
\bauthor{\bsnm{Georgiev}, \binits{P.}}, \betal:
\batitle{Grandmaster level in starcraft ii using multi-agent reinforcement
  learning}.
\bjtitle{Nature}
\bvolume{575}(\bissue{7782}),
\bfpage{350}--\blpage{354}
(\byear{2019})
\end{barticle}
\endbibitem

\bibitem[\protect\citeauthoryear{Ceron and Castro}{2021}]{ceron2021revisiting}
\begin{bchapter}
\bauthor{\bsnm{Ceron}, \binits{J.S.O.}},
\bauthor{\bsnm{Castro}, \binits{P.S.}}:
\bctitle{Revisiting rainbow: Promoting more insightful and inclusive deep
  reinforcement learning research}.
In: \bbtitle{International Conference on Machine Learning},
pp. \bfpage{1373}--\blpage{1383}
(\byear{2021}).
\bcomment{PMLR}
\end{bchapter}
\endbibitem

\bibitem[\protect\citeauthoryear{Fern{\'a}ndez and
  Veloso}{2006}]{fernandez2006probabilistic}
\begin{bchapter}
\bauthor{\bsnm{Fern{\'a}ndez}, \binits{F.}},
\bauthor{\bsnm{Veloso}, \binits{M.}}:
\bctitle{Probabilistic policy reuse in a reinforcement learning agent}.
In: \bbtitle{Proceedings of the Fifth International Joint Conference on
  Autonomous Agents and Multiagent Systems},
pp. \bfpage{720}--\blpage{727}
(\byear{2006})
\end{bchapter}
\endbibitem

\bibitem[\protect\citeauthoryear{Barreto et~al.}{2018}]{barreto2018transfer}
\begin{bchapter}
\bauthor{\bsnm{Barreto}, \binits{A.}},
\bauthor{\bsnm{Borsa}, \binits{D.}},
\bauthor{\bsnm{Quan}, \binits{J.}},
\bauthor{\bsnm{Schaul}, \binits{T.}},
\bauthor{\bsnm{Silver}, \binits{D.}},
\bauthor{\bsnm{Hessel}, \binits{M.}},
\bauthor{\bsnm{Mankowitz}, \binits{D.}},
\bauthor{\bsnm{Zidek}, \binits{A.}},
\bauthor{\bsnm{Munos}, \binits{R.}}:
\bctitle{Transfer in deep reinforcement learning using successor features and
  generalised policy improvement}.
In: \bbtitle{International Conference on Machine Learning},
pp. \bfpage{501}--\blpage{510}
(\byear{2018}).
\bcomment{PMLR}
\end{bchapter}
\endbibitem

\bibitem[\protect\citeauthoryear{Li et~al.}{2019}]{li2019hierarchical}
\begin{botherref}
\oauthor{\bsnm{Li}, \binits{S.}},
\oauthor{\bsnm{Wang}, \binits{R.}},
\oauthor{\bsnm{Tang}, \binits{M.}},
\oauthor{\bsnm{Zhang}, \binits{C.}}:
Hierarchical reinforcement learning with advantage-based auxiliary rewards.
Advances in Neural Information Processing Systems
\textbf{32}
(2019)
\end{botherref}
\endbibitem

\bibitem[\protect\citeauthoryear{Yang et~al.}{2020}]{yang2020efficient}
\begin{bchapter}
\bauthor{\bsnm{Yang}, \binits{T.}},
\bauthor{\bsnm{Hao}, \binits{J.}},
\bauthor{\bsnm{Meng}, \binits{Z.}},
\bauthor{\bsnm{Zhang}, \binits{Z.}},
\bauthor{\bsnm{Hu}, \binits{Y.}},
\bauthor{\bsnm{Chen}, \binits{Y.}},
\bauthor{\bsnm{Fan}, \binits{C.}},
\bauthor{\bsnm{Wang}, \binits{W.}},
\bauthor{\bsnm{Liu}, \binits{W.}},
\bauthor{\bsnm{Wang}, \binits{Z.}},
\bauthor{\bsnm{Peng}, \binits{J.}}:
\bctitle{Efficient deep reinforcement learning via adaptive policy transfer}.
In: \bbtitle{Proceedings of the Twenty-Ninth International Joint Conference on
  Artificial Intelligence, {IJCAI-20}},
pp. \bfpage{3094}--\blpage{3100}
(\byear{2020})
\end{bchapter}
\endbibitem

\bibitem[\protect\citeauthoryear{Zhang et~al.}{2022}]{zhang2022cup}
\begin{bchapter}
\bauthor{\bsnm{Zhang}, \binits{J.}},
\bauthor{\bsnm{Li}, \binits{S.}},
\bauthor{\bsnm{Zhang}, \binits{C.}}:
\bctitle{Cup: Critic-guided policy reuse}.
In: \bbtitle{Advances in Neural Information Processing Systems}
(\byear{2022})
\end{bchapter}
\endbibitem

\bibitem[\protect\citeauthoryear{Li et~al.}{2018}]{li2018context}
\begin{botherref}
\oauthor{\bsnm{Li}, \binits{S.}},
\oauthor{\bsnm{Gu}, \binits{F.}},
\oauthor{\bsnm{Zhu}, \binits{G.}},
\oauthor{\bsnm{Zhang}, \binits{C.}}:
Context-aware policy reuse.
arXiv preprint arXiv:1806.03793
(2018)
\end{botherref}
\endbibitem

\bibitem[\protect\citeauthoryear{Kurenkov et~al.}{2020}]{acteach}
\begin{bchapter}
\bauthor{\bsnm{Kurenkov}, \binits{A.}},
\bauthor{\bsnm{Mandlekar}, \binits{A.}},
\bauthor{\bsnm{Martin-Martin}, \binits{R.}},
\bauthor{\bsnm{Savarese}, \binits{S.}},
\bauthor{\bsnm{Garg}, \binits{A.}}:
\bctitle{Ac-teach: A bayesian actor-critic method for policy learning with an
  ensemble of suboptimal teachers}.
In: \bbtitle{Conference on Robot Learning},
pp. \bfpage{717}--\blpage{734}
(\byear{2020}).
\bcomment{PMLR}
\end{bchapter}
\endbibitem

\bibitem[\protect\citeauthoryear{Barreto et~al.}{2017}]{barreto2017successor}
\begin{botherref}
\oauthor{\bsnm{Barreto}, \binits{A.}},
\oauthor{\bsnm{Dabney}, \binits{W.}},
\oauthor{\bsnm{Munos}, \binits{R.}},
\oauthor{\bsnm{Hunt}, \binits{J.J.}},
\oauthor{\bsnm{Schaul}, \binits{T.}},
\oauthor{\bsnm{Hasselt}, \binits{H.P.}},
\oauthor{\bsnm{Silver}, \binits{D.}}:
Successor features for transfer in reinforcement learning.
Advances in neural information processing systems
\textbf{30}
(2017)
\end{botherref}
\endbibitem

\bibitem[\protect\citeauthoryear{Cheng et~al.}{2020}]{cheng2020policy}
\begin{barticle}
\bauthor{\bsnm{Cheng}, \binits{C.-A.}},
\bauthor{\bsnm{Kolobov}, \binits{A.}},
\bauthor{\bsnm{Agarwal}, \binits{A.}}:
\batitle{Policy improvement via imitation of multiple oracles}.
\bjtitle{Advances in Neural Information Processing Systems}
\bvolume{33},
\bfpage{5587}--\blpage{5598}
(\byear{2020})
\end{barticle}
\endbibitem

\bibitem[\protect\citeauthoryear{Pateria
  et~al.}{2021}]{pateria2021hierarchical}
\begin{barticle}
\bauthor{\bsnm{Pateria}, \binits{S.}},
\bauthor{\bsnm{Subagdja}, \binits{B.}},
\bauthor{\bsnm{Tan}, \binits{A.-h.}},
\bauthor{\bsnm{Quek}, \binits{C.}}:
\batitle{Hierarchical reinforcement learning: A comprehensive survey}.
\bjtitle{ACM Computing Surveys (CSUR)}
\bvolume{54}(\bissue{5}),
\bfpage{1}--\blpage{35}
(\byear{2021})
\end{barticle}
\endbibitem

\bibitem[\protect\citeauthoryear{Lillicrap
  et~al.}{2016}]{lillicrap2016continuous}
\begin{bchapter}
\bauthor{\bsnm{Lillicrap}, \binits{T.P.}},
\bauthor{\bsnm{Hunt}, \binits{J.J.}},
\bauthor{\bsnm{Pritzel}, \binits{A.}},
\bauthor{\bsnm{Heess}, \binits{N.}},
\bauthor{\bsnm{Erez}, \binits{T.}},
\bauthor{\bsnm{Tassa}, \binits{Y.}},
\bauthor{\bsnm{Silver}, \binits{D.}},
\bauthor{\bsnm{Wierstra}, \binits{D.}}:
\bctitle{Continuous control with deep reinforcement learning.}
In: \bbtitle{ICLR (Poster)}
(\byear{2016})
\end{bchapter}
\endbibitem

\bibitem[\protect\citeauthoryear{Fujimoto
  et~al.}{2018}]{fujimoto2018addressing}
\begin{bchapter}
\bauthor{\bsnm{Fujimoto}, \binits{S.}},
\bauthor{\bsnm{Hoof}, \binits{H.}},
\bauthor{\bsnm{Meger}, \binits{D.}}:
\bctitle{Addressing function approximation error in actor-critic methods}.
In: \bbtitle{International Conference on Machine Learning},
pp. \bfpage{1587}--\blpage{1596}
(\byear{2018}).
\bcomment{PMLR}
\end{bchapter}
\endbibitem

\bibitem[\protect\citeauthoryear{Haarnoja et~al.}{2018}]{haarnoja2018soft2}
\begin{botherref}
\oauthor{\bsnm{Haarnoja}, \binits{T.}},
\oauthor{\bsnm{Zhou}, \binits{A.}},
\oauthor{\bsnm{Hartikainen}, \binits{K.}},
\oauthor{\bsnm{Tucker}, \binits{G.}},
\oauthor{\bsnm{Ha}, \binits{S.}},
\oauthor{\bsnm{Tan}, \binits{J.}},
\oauthor{\bsnm{Kumar}, \binits{V.}},
\oauthor{\bsnm{Zhu}, \binits{H.}},
\oauthor{\bsnm{Gupta}, \binits{A.}},
\oauthor{\bsnm{Abbeel}, \binits{P.}}, et al.:
Soft actor-critic algorithms and applications.
arXiv preprint arXiv:1812.05905
(2018)
\end{botherref}
\endbibitem

\bibitem[\protect\citeauthoryear{Yu et~al.}{2020}]{yu2020meta}
\begin{bchapter}
\bauthor{\bsnm{Yu}, \binits{T.}},
\bauthor{\bsnm{Quillen}, \binits{D.}},
\bauthor{\bsnm{He}, \binits{Z.}},
\bauthor{\bsnm{Julian}, \binits{R.}},
\bauthor{\bsnm{Hausman}, \binits{K.}},
\bauthor{\bsnm{Finn}, \binits{C.}},
\bauthor{\bsnm{Levine}, \binits{S.}}:
\bctitle{Meta-world: A benchmark and evaluation for multi-task and meta
  reinforcement learning}.
In: \bbtitle{Conference on Robot Learning},
pp. \bfpage{1094}--\blpage{1100}
(\byear{2020}).
\bcomment{PMLR}
\end{bchapter}
\endbibitem

\bibitem[\protect\citeauthoryear{Zhu et~al.}{2020}]{zhu2020transfer}
\begin{botherref}
\oauthor{\bsnm{Zhu}, \binits{Z.}},
\oauthor{\bsnm{Lin}, \binits{K.}},
\oauthor{\bsnm{Zhou}, \binits{J.}}:
Transfer learning in deep reinforcement learning: A survey.
arXiv preprint arXiv:2009.07888
(2020)
\end{botherref}
\endbibitem

\bibitem[\protect\citeauthoryear{Parisotto et~al.}{2015}]{parisotto2015actor}
\begin{botherref}
\oauthor{\bsnm{Parisotto}, \binits{E.}},
\oauthor{\bsnm{Ba}, \binits{J.L.}},
\oauthor{\bsnm{Salakhutdinov}, \binits{R.}}:
Actor-mimic: Deep multitask and transfer reinforcement learning.
arXiv preprint arXiv:1511.06342
(2015)
\end{botherref}
\endbibitem

\bibitem[\protect\citeauthoryear{Hou et~al.}{2017}]{hou2017evolutionary}
\begin{barticle}
\bauthor{\bsnm{Hou}, \binits{Y.}},
\bauthor{\bsnm{Ong}, \binits{Y.-S.}},
\bauthor{\bsnm{Feng}, \binits{L.}},
\bauthor{\bsnm{Zurada}, \binits{J.M.}}:
\batitle{An evolutionary transfer reinforcement learning framework for
  multiagent systems}.
\bjtitle{IEEE Transactions on Evolutionary Computation}
\bvolume{21}(\bissue{4}),
\bfpage{601}--\blpage{615}
(\byear{2017})
\end{barticle}
\endbibitem

\bibitem[\protect\citeauthoryear{Laroche and
  Barlier}{2017}]{laroche2017transfer}
\begin{bchapter}
\bauthor{\bsnm{Laroche}, \binits{R.}},
\bauthor{\bsnm{Barlier}, \binits{M.}}:
\bctitle{Transfer reinforcement learning with shared dynamics}.
In: \bbtitle{Thirty-First AAAI Conference on Artificial Intelligence}
(\byear{2017})
\end{bchapter}
\endbibitem

\bibitem[\protect\citeauthoryear{Lehnert and Littman}{2020}]{JMLRv2119-060}
\begin{barticle}
\bauthor{\bsnm{Lehnert}, \binits{L.}},
\bauthor{\bsnm{Littman}, \binits{M.L.}}:
\batitle{Successor features combine elements of model-free and model-based
  reinforcement learning}.
\bjtitle{Journal of Machine Learning Research}
\bvolume{21}(\bissue{196}),
\bfpage{1}--\blpage{53}
(\byear{2020})
\end{barticle}
\endbibitem

\bibitem[\protect\citeauthoryear{Barekatain
  et~al.}{2020}]{barekatain2021multipolar}
\begin{bchapter}
\bauthor{\bsnm{Barekatain}, \binits{M.}},
\bauthor{\bsnm{Yonetani}, \binits{R.}},
\bauthor{\bsnm{Hamaya}, \binits{M.}}:
\bctitle{Multipolar: multi-source policy aggregation for transfer reinforcement
  learning between diverse environmental dynamics}.
In: \bbtitle{Proceedings of the Twenty-Ninth International Conference on
  International Joint Conferences on Artificial Intelligence},
pp. \bfpage{3108}--\blpage{3116}
(\byear{2020})
\end{bchapter}
\endbibitem

\bibitem[\protect\citeauthoryear{Li and Zhang}{2018}]{li2018optimal}
\begin{bchapter}
\bauthor{\bsnm{Li}, \binits{S.}},
\bauthor{\bsnm{Zhang}, \binits{C.}}:
\bctitle{An optimal online method of selecting source policies for
  reinforcement learning}.
In: \bbtitle{Proceedings of the AAAI Conference on Artificial Intelligence},
vol. \bseriesno{32}
(\byear{2018})
\end{bchapter}
\endbibitem

\bibitem[\protect\citeauthoryear{Gimelfarb
  et~al.}{2021}]{gimelfarb2021contextual}
\begin{bchapter}
\bauthor{\bsnm{Gimelfarb}, \binits{M.}},
\bauthor{\bsnm{Sanner}, \binits{S.}},
\bauthor{\bsnm{Lee}, \binits{C.-G.}}:
\bctitle{Contextual policy transfer in reinforcement learning domains via deep
  mixtures-of-experts}.
In: \bbtitle{Uncertainty in Artificial Intelligence},
pp. \bfpage{1787}--\blpage{1797}
(\byear{2021}).
\bcomment{PMLR}
\end{bchapter}
\endbibitem

\bibitem[\protect\citeauthoryear{Yang et~al.}{2021}]{yang2021hierarchical}
\begin{barticle}
\bauthor{\bsnm{Yang}, \binits{X.}},
\bauthor{\bsnm{Ji}, \binits{Z.}},
\bauthor{\bsnm{Wu}, \binits{J.}},
\bauthor{\bsnm{Lai}, \binits{Y.-K.}},
\bauthor{\bsnm{Wei}, \binits{C.}},
\bauthor{\bsnm{Liu}, \binits{G.}},
\bauthor{\bsnm{Setchi}, \binits{R.}}:
\batitle{Hierarchical reinforcement learning with universal policies for
  multistep robotic manipulation}.
\bjtitle{IEEE Transactions on Neural Networks and Learning Systems}
\bvolume{33}(\bissue{9}),
\bfpage{4727}--\blpage{4741}
(\byear{2021})
\end{barticle}
\endbibitem

\bibitem[\protect\citeauthoryear{Rusu et~al.}{2016}]{rusu2016progressive}
\begin{botherref}
\oauthor{\bsnm{Rusu}, \binits{A.A.}},
\oauthor{\bsnm{Rabinowitz}, \binits{N.C.}},
\oauthor{\bsnm{Desjardins}, \binits{G.}},
\oauthor{\bsnm{Soyer}, \binits{H.}},
\oauthor{\bsnm{Kirkpatrick}, \binits{J.}},
\oauthor{\bsnm{Kavukcuoglu}, \binits{K.}},
\oauthor{\bsnm{Pascanu}, \binits{R.}},
\oauthor{\bsnm{Hadsell}, \binits{R.}}:
Progressive neural networks.
arXiv preprint arXiv:1606.04671
(2016)
\end{botherref}
\endbibitem

\bibitem[\protect\citeauthoryear{Berseth et~al.}{2018}]{berseth2018progressive}
\begin{botherref}
\oauthor{\bsnm{Berseth}, \binits{G.}},
\oauthor{\bsnm{Xie}, \binits{C.}},
\oauthor{\bsnm{Cernek}, \binits{P.}},
\oauthor{\bsnm{Panne}, \binits{M.}}:
Progressive reinforcement learning with distillation for multi-skilled motion
  control.
arXiv preprint arXiv:1802.04765
(2018)
\end{botherref}
\endbibitem

\bibitem[\protect\citeauthoryear{Schwarz et~al.}{2018}]{schwarz2018progress}
\begin{bchapter}
\bauthor{\bsnm{Schwarz}, \binits{J.}},
\bauthor{\bsnm{Czarnecki}, \binits{W.}},
\bauthor{\bsnm{Luketina}, \binits{J.}},
\bauthor{\bsnm{Grabska-Barwinska}, \binits{A.}},
\bauthor{\bsnm{Teh}, \binits{Y.W.}},
\bauthor{\bsnm{Pascanu}, \binits{R.}},
\bauthor{\bsnm{Hadsell}, \binits{R.}}:
\bctitle{Progress \& compress: A scalable framework for continual learning}.
In: \bbtitle{International Conference on Machine Learning},
pp. \bfpage{4528}--\blpage{4537}
(\byear{2018}).
\bcomment{PMLR}
\end{bchapter}
\endbibitem

\bibitem[\protect\citeauthoryear{Mallya and Lazebnik}{2018}]{mallya2018packnet}
\begin{bchapter}
\bauthor{\bsnm{Mallya}, \binits{A.}},
\bauthor{\bsnm{Lazebnik}, \binits{S.}}:
\bctitle{Packnet: Adding multiple tasks to a single network by iterative
  pruning}.
In: \bbtitle{Proceedings of the IEEE Conference on Computer Vision and Pattern
  Recognition},
pp. \bfpage{7765}--\blpage{7773}
(\byear{2018})
\end{bchapter}
\endbibitem

\bibitem[\protect\citeauthoryear{Teh et~al.}{2017}]{teh2017distral}
\begin{botherref}
\oauthor{\bsnm{Teh}, \binits{Y.}},
\oauthor{\bsnm{Bapst}, \binits{V.}},
\oauthor{\bsnm{Czarnecki}, \binits{W.M.}},
\oauthor{\bsnm{Quan}, \binits{J.}},
\oauthor{\bsnm{Kirkpatrick}, \binits{J.}},
\oauthor{\bsnm{Hadsell}, \binits{R.}},
\oauthor{\bsnm{Heess}, \binits{N.}},
\oauthor{\bsnm{Pascanu}, \binits{R.}}:
Distral: Robust multitask reinforcement learning.
Advances in neural information processing systems
\textbf{30}
(2017)
\end{botherref}
\endbibitem

\bibitem[\protect\citeauthoryear{Haarnoja et~al.}{2018}]{haarnoja2018soft}
\begin{bchapter}
\bauthor{\bsnm{Haarnoja}, \binits{T.}},
\bauthor{\bsnm{Zhou}, \binits{A.}},
\bauthor{\bsnm{Abbeel}, \binits{P.}},
\bauthor{\bsnm{Levine}, \binits{S.}}:
\bctitle{Soft actor-critic: Off-policy maximum entropy deep reinforcement
  learning with a stochastic actor}.
In: \bbtitle{International Conference on Machine Learning},
pp. \bfpage{1861}--\blpage{1870}
(\byear{2018}).
\bcomment{PMLR}
\end{bchapter}
\endbibitem

\bibitem[\protect\citeauthoryear{Lan et~al.}{2020}]{lan2020maxmin}
\begin{botherref}
\oauthor{\bsnm{Lan}, \binits{Q.}},
\oauthor{\bsnm{Pan}, \binits{Y.}},
\oauthor{\bsnm{Fyshe}, \binits{A.}},
\oauthor{\bsnm{White}, \binits{M.}}:
Maxmin q-learning: Controlling the estimation bias of q-learning.
arXiv preprint arXiv:2002.06487
(2020)
\end{botherref}
\endbibitem

\bibitem[\protect\citeauthoryear{Kuznetsov
  et~al.}{2020}]{kuznetsov2020controlling}
\begin{bchapter}
\bauthor{\bsnm{Kuznetsov}, \binits{A.}},
\bauthor{\bsnm{Shvechikov}, \binits{P.}},
\bauthor{\bsnm{Grishin}, \binits{A.}},
\bauthor{\bsnm{Vetrov}, \binits{D.}}:
\bctitle{Controlling overestimation bias with truncated mixture of continuous
  distributional quantile critics}.
In: \bbtitle{International Conference on Machine Learning},
pp. \bfpage{5556}--\blpage{5566}
(\byear{2020}).
\bcomment{PMLR}
\end{bchapter}
\endbibitem

\bibitem[\protect\citeauthoryear{Zhang and Sutton}{2017}]{zhang2017deeper}
\begin{botherref}
\oauthor{\bsnm{Zhang}, \binits{S.}},
\oauthor{\bsnm{Sutton}, \binits{R.S.}}:
A deeper look at experience replay.
arXiv preprint arXiv:1712.01275
(2017)
\end{botherref}
\endbibitem

\bibitem[\protect\citeauthoryear{Fedus et~al.}{2020}]{fedus2020revisiting}
\begin{bchapter}
\bauthor{\bsnm{Fedus}, \binits{W.}},
\bauthor{\bsnm{Ramachandran}, \binits{P.}},
\bauthor{\bsnm{Agarwal}, \binits{R.}},
\bauthor{\bsnm{Bengio}, \binits{Y.}},
\bauthor{\bsnm{Larochelle}, \binits{H.}},
\bauthor{\bsnm{Rowland}, \binits{M.}},
\bauthor{\bsnm{Dabney}, \binits{W.}}:
\bctitle{Revisiting fundamentals of experience replay}.
In: \bbtitle{International Conference on Machine Learning},
pp. \bfpage{3061}--\blpage{3071}
(\byear{2020}).
\bcomment{PMLR}
\end{bchapter}
\endbibitem

\bibitem[\protect\citeauthoryear{Khetarpal et~al.}{2020}]{khetarpal2020towards}
\begin{botherref}
\oauthor{\bsnm{Khetarpal}, \binits{K.}},
\oauthor{\bsnm{Riemer}, \binits{M.}},
\oauthor{\bsnm{Rish}, \binits{I.}},
\oauthor{\bsnm{Precup}, \binits{D.}}:
Towards continual reinforcement learning: A review and perspectives.
arXiv preprint arXiv:2012.13490
(2020)
\end{botherref}
\endbibitem

\bibitem[\protect\citeauthoryear{Wo{l}czyk et~al.}{2021}]{wolczyk2021continual}
\begin{barticle}
\bauthor{\bsnm{Wo{l}czyk}, \binits{M.}},
\bauthor{\bsnm{Zaj{k{a}}c}, \binits{M.}},
\bauthor{\bsnm{Pascanu}, \binits{R.}},
\bauthor{\bsnm{Kuci{n}ski}, \binits{L.}},
\bauthor{\bsnm{Mi{l}o{s}}, \binits{P.}}:
\batitle{Continual world: A robotic benchmark for continual reinforcement
  learning}.
\bjtitle{Advances in Neural Information Processing Systems}
\bvolume{34},
\bfpage{28496}--\blpage{28510}
(\byear{2021})
\end{barticle}
\endbibitem

\bibitem[\protect\citeauthoryear{Yang et~al.}{2020}]{yang2020multi}
\begin{barticle}
\bauthor{\bsnm{Yang}, \binits{R.}},
\bauthor{\bsnm{Xu}, \binits{H.}},
\bauthor{\bsnm{Wu}, \binits{Y.}},
\bauthor{\bsnm{Wang}, \binits{X.}}:
\batitle{Multi-task reinforcement learning with soft modularization}.
\bjtitle{Advances in Neural Information Processing Systems}
\bvolume{33},
\bfpage{4767}--\blpage{4777}
(\byear{2020})
\end{barticle}
\endbibitem

\bibitem[\protect\citeauthoryear{Sodhani et~al.}{2021}]{sodhani2021multi}
\begin{bchapter}
\bauthor{\bsnm{Sodhani}, \binits{S.}},
\bauthor{\bsnm{Zhang}, \binits{A.}},
\bauthor{\bsnm{Pineau}, \binits{J.}}:
\bctitle{Multi-task reinforcement learning with context-based representations}.
In: \bbtitle{International Conference on Machine Learning},
pp. \bfpage{9767}--\blpage{9779}
(\byear{2021}).
\bcomment{PMLR}
\end{bchapter}
\endbibitem

\bibitem[\protect\citeauthoryear{Wan et~al.}{2020}]{wan2020mutual}
\begin{botherref}
\oauthor{\bsnm{Wan}, \binits{M.}},
\oauthor{\bsnm{Gangwani}, \binits{T.}},
\oauthor{\bsnm{Peng}, \binits{J.}}:
Mutual information based knowledge transfer under state-action dimension
  mismatch.
arXiv preprint arXiv:2006.07041
(2020)
\end{botherref}
\endbibitem

\bibitem[\protect\citeauthoryear{Zhang et~al.}{2020}]{zhang2020learning}
\begin{botherref}
\oauthor{\bsnm{Zhang}, \binits{Q.}},
\oauthor{\bsnm{Xiao}, \binits{T.}},
\oauthor{\bsnm{Efros}, \binits{A.A.}},
\oauthor{\bsnm{Pinto}, \binits{L.}},
\oauthor{\bsnm{Wang}, \binits{X.}}:
Learning cross-domain correspondence for control with dynamics
  cycle-consistency.
arXiv preprint arXiv:2012.09811
(2020)
\end{botherref}
\endbibitem

\bibitem[\protect\citeauthoryear{Heng et~al.}{2022}]{heng2022cross}
\begin{bchapter}
\bauthor{\bsnm{Heng}, \binits{Y.}},
\bauthor{\bsnm{Yang}, \binits{T.}},
\bauthor{\bsnm{ZHENG}, \binits{Y.}},
\bauthor{\bsnm{Jianye}, \binits{H.}},
\bauthor{\bsnm{Taylor}, \binits{M.E.}}:
\bctitle{Cross-domain adaptive transfer reinforcement learning based on
  state-action correspondence}.
In: \bbtitle{The 38th Conference on Uncertainty in Artificial Intelligence}
(\byear{2022})
\end{bchapter}
\endbibitem

\bibitem[\protect\citeauthoryear{van~der Pol et~al.}{2020a}]{van2020mdp}
\begin{barticle}
\bauthor{\bsnm{Pol}, \binits{E.}},
\bauthor{\bsnm{Worrall}, \binits{D.}},
\bauthor{\bsnm{Hoof}, \binits{H.}},
\bauthor{\bsnm{Oliehoek}, \binits{F.}},
\bauthor{\bsnm{Welling}, \binits{M.}}:
\batitle{Mdp homomorphic networks: Group symmetries in reinforcement learning}.
\bjtitle{Advances in Neural Information Processing Systems}
\bvolume{33},
\bfpage{4199}--\blpage{4210}
(\byear{2020})
\end{barticle}
\endbibitem

\bibitem[\protect\citeauthoryear{van~der Pol et~al.}{2020b}]{van2020plannable}
\begin{botherref}
\oauthor{\bsnm{Pol}, \binits{E.}},
\oauthor{\bsnm{Kipf}, \binits{T.}},
\oauthor{\bsnm{Oliehoek}, \binits{F.A.}},
\oauthor{\bsnm{Welling}, \binits{M.}}:
Plannable approximations to mdp homomorphisms: Equivariance under actions.
arXiv preprint arXiv:2002.11963
(2020)
\end{botherref}
\endbibitem

\bibitem[\protect\citeauthoryear{Fedotov et~al.}{2003}]{fedotov2003refinements}
\begin{barticle}
\bauthor{\bsnm{Fedotov}, \binits{A.A.}},
\bauthor{\bsnm{Harremo{\"e}s}, \binits{P.}},
\bauthor{\bsnm{Topsoe}, \binits{F.}}:
\batitle{Refinements of pinsker's inequality}.
\bjtitle{IEEE Transactions on Information Theory}
\bvolume{49}(\bissue{6}),
\bfpage{1491}--\blpage{1498}
(\byear{2003})
\end{barticle}
\endbibitem

\bibitem[\protect\citeauthoryear{Kakade and
  Langford}{2002}]{kakade2002approximately}
\begin{bchapter}
\bauthor{\bsnm{Kakade}, \binits{S.}},
\bauthor{\bsnm{Langford}, \binits{J.}}:
\bctitle{Approximately optimal approximate reinforcement learning}.
In: \bbtitle{In Proc. 19th International Conference on Machine Learning}
(\byear{2002}).
\bcomment{Citeseer}
\end{bchapter}
\endbibitem

\end{thebibliography}

\end{document}